\newif\ifmulticol	\multicoltrue
\newif\ifshowgit	\showgittrue		% switches footer on/off
\newif\ifgitlocal	\gitlocaltrue		% use local file gitHeadLocal.gin
\newif\ifbiblatex	\biblatexfalse		% defaults to bibtex if false
\newif\ifbibnum		\bibnumtrue 		% num => superscripts, otherwise auth date
\newif\ifbibsort	\bibsortfalse		% biblatex num sort in order of occurrence
\newif\iflineno		\linenofalse
\newif\iftoc		\toctrue

\newif\iflucida		\lucidafalse
\newif\ifcm			\cmfalse
\newif\ifcharter	\chartertrue		% use for arXiv, requires xelatex
\newif\ifcharterotf	\charterotffalse	% requires xelatex

\newcommand*{\secnumdepth}{0}			% sets secnumdepth in \mymaketitle
\newcommand*{\tocdepth}{2}				% sets tocdepth in \mymaketitle
\newcommand*{\reftitle}{References}		% title for references section
\newcommand*{\mytitleskip}{-0.2in}		% change skip after title

\newcommand*{\mydocfontsize}{\ifcharter11pt\else10pt\fi}
\newcommand*{\setcol}{\ifmulticol twocolumn\else onecolumn\fi}

\documentclass[\mydocfontsize,\setcol]{article}

\newcommand*{\pdfauthor}{Steven A.~Frank}
\newcommand*{\pdftitle}{The Price equation reveals a universal force-metric-bias law of algorithmic learning and natural selection}

\input learn.sty
\newcommand*{\bth}{\bm{\Gth}}
\newcommand*{\btht}{\skew{2}\tilde{\bm{\Gth}}}

\newcommand*{\bthb}{\bar{\bm{\Gth}}}

\newcommand*{\Gthb}{\bar{\Gth}}
\newcommand*{\bthbg}{\bar{\bm{\Gth}}_g}
\newcommand*{\bthbgj}{{\bm{\Gth}}_{j|g}}

\newcommand*{\br}{\bmr{r}}
\newcommand*{\bx}{\bmr{x}}
\newcommand*{\bxt}{\tilde{\bx}}
\newcommand*{\bxh}{\hat{\bx}}
\newcommand*{\by}{\bmr{y}}
\newcommand*{\bb}{\bmr{b}}
\newcommand*{\bbg}{\bmr{b}_g}
\newcommand*{\bbgt}{\skew{-4}\tilde{\bb}_g}
\newcommand*{\bff}{\bmr{f}}
\newcommand*{\bfB}{\bmr{f}_B}
\newcommand*{\bfg}{\bmr{f}_g}
\newcommand*{\bg}{\bmr{g}}
\newcommand*{\bv}{\bmr{v}}

\newcommand*{\ba}{\bmr{a}}

\newcommand*{\bC}{\bmr{C}}

\newcommand*{\bD}{\bmr{D}}
\newcommand*{\bF}{\bmr{F}}

\newcommand*{\bG}{\bmr{G}}
\newcommand*{\bGi}{\bmr{G}^{-1}}
\newcommand*{\bH}{\bmr{H}}
\newcommand*{\bHi}{\bH^{-1}}
\newcommand*{\bHt}{\bmr{\tilde{H}}}
\newcommand*{\bHit}{\bmr{\tilde{H}}^{-1}}
\newcommand*{\bI}{\bmr{I}}
\newcommand*{\bJ}{\bmr{J}}
\newcommand*{\bK}{\bmr{K}}
\newcommand*{\bM}{\bmr{M}}
\newcommand*{\bMi}{\bM^{-1}}
\newcommand*{\bMB}{\bmr{M}_B}
\newcommand*{\bMg}{\bmr{M}_g}
\newcommand*{\bP}{\bmr{P}}
\newcommand*{\bQ}{\bmr{Q}}
\newcommand*{\bR}{\bmr{R}}
\newcommand*{\bS}{\bmr{S}}
\newcommand*{\bX}{\bmr{X}}

\newcommand*{\bxi}{\bm{\xi}}
\newcommand*{\bGe}{\bm{\Ge}}
\newcommand*{\bGa}{\bm{\Ga}}
\newcommand*{\bGb}{\bm{\Gb}}

\newcommand*{\bGf}{\bm{\Gf}}
\newcommand*{\bGg}{\bm{\Gg}}

\newcommand*{\nsub}[1]{\nabla\mskip-3mu_{#1}\mskip1mu}

\newcommand*{\bq}{\bmr{q}}
\newcommand*{\dbq}{\GD\bmr{q}}
\newcommand*{\ddbq}{\dd\bmr{q}}
\newcommand*{\qhat}{\hat{q}}
\newcommand*{\bqhat}{\hat{\bq}}

\newcommand{\LL}{\mathcal{L}}
\newcommand*{\U}{U}
\newcommand*{\F}{\mathcal{F}}
\newcommand*{\J}{\mathcal{J}}
\newcommand*{\N}{\mathcal{N}}
\newcommand{\KL}[2]{\D\left(#1||#2\right)}
\newcommand{\D}{\mathcal D}
\newcommand{\bL}{\mathbf{L}}
\newcommand{\bLt}{\skew{-3}\tilde{\mathbf{L}}}
\newcommand{\Lt}{\skew{-0.4}\tilde{L}}

\newcommand*{\cov}{\mathrm{Cov}}
\newcommand*{\var}{\mathrm{Var}}
\newcommand*{\reg}{\mathrm{Reg}}
\newcommand*{\bmm}{\bmr{m}}
\newcommand*{\bw}{\bmr{w}}
\newcommand*{\wbar}{\bar{w}}
\newcommand*{\wbg}{\bar{w}_g}
\newcommand*{\wbgj}{{w}_{j|g}}
\newcommand*{\mbar}{\bar{m}}
\newcommand*{\gbar}{\bar{\bg}}

\newcommand*{\df}{\GD_{\bff}\mskip2mu}

\newcommand*{\db}{\GD_{\bb}\mskip2mu}
\newcommand*{\dxi}{\GD_{\bxi}\mskip2mu}

\newcommand{\dbz}{\GD\mathbf{z}}
\newcommand{\bz}{\mathbf{z}}
\newcommand{\zhat}{\hat{z}}
\newcommand{\zbar}{\bar{z}}

\newcommand*{\Ga}{\alpha}
\newcommand*{\Gb}{\beta}
\newcommand*{\Gd}{\delta}
\newcommand*{\GD}{\Delta}
\newcommand*{\Ge}{\epsilon}
\newcommand*{\Gg}{\gamma}

\newcommand*{\Gl}{\lambda}

\newcommand*{\Gm}{\mu}

\newcommand*{\Gs}{\sigma}
\newcommand*{\Gt}{\tau}
\newcommand*{\Gth}{\theta}
\newcommand*{\Gf}{\phi}

\usepackage{bm}
\newcommand{\bmr}[1]{\bm{\mathrm{#1}}}

\DeclarePairedDelimiter\abs{\lvert}{\rvert}
\DeclarePairedDelimiter\norm{\lVert}{\rVert}
\DeclarePairedDelimiter\angb{\langle}{\rangle}
\DeclarePairedDelimiter\lrb{\lbrack}{\rbrack}
\DeclarePairedDelimiter\lr{\lparen}{\rparen}
\DeclarePairedDelimiter\lrbr{\lbrace}{\rbrace}

\makeatletter
\let\oldabs\abs \def\abs{\@ifstar{\oldabs}{\oldabs*}}
\let\oldnorm\norm \def\norm{\@ifstar{\oldnorm}{\oldnorm*}}
\let\oldangb\angb \def\angb{\@ifstar{\oldangb}{\oldangb*}}
\let\oldlrb\lrb \def\lrb{\@ifstar{\oldlrb}{\oldlrb*}}
\let\oldlr\lr \def\lr{\@ifstar{\oldlr}{\oldlr*}}
\let\oldlrbr\lrbr \def\lrbr{\@ifstar{\oldlrbr}{\oldlrbr*}}
\makeatother

\DeclareMathOperator{\E}{E}
\newcommand*{\dd}{\textrm{d}}
\newcommand*{\Eq}[1]{eqn~\ref{eq:#1}}

\newcommand*{\prt}{\partial}

\newcommand*{\Figure}[1]{Figure~\ref{fig:#1}}

\newcommand*{\boldrule}{\hrule height 1.2pt}
\newcommand*{\noterule}{\medskip\boldrule\medskip}	% for notes

\hypersetup{linkcolor={black}, urlcolor={black}}

\newcommand*{\mytitle}{\pdftitle}

\newcommand*{\myauthor}{Steven A.\ Frank}
\newcommand*{\myaddress}{Department of Ecology and Evolutionary Biology, University of California, Irvine, CA 92697--2525, USA}
\newcommand*{\mycontact}{\href{https://stevefrank.org}{https://stevefrank.org}}

\newcommand*{\mykeywords}{Price equation; force-metric-bias; Bayesian updating; momentum-based optimization; information geometry; natural selection; stochastic gradient descent\vskip30pt}
\setabstract{-0.07}{\iftoc-2.4\else1.22\fi}{%
Diverse learning algorithms, optimization methods, and natural selection share a common mathematical structure, despite their apparent differences. Here I show that a simple notational partitioning of change by the Price equation reveals a universal force-metric-bias (FMB) law: $\GD\bth = \bM\,\bff + \bb + \bxi$. The force $\bff$ drives improvement in parameters, $\GD\bth$, in proportion to the slope of performance with respect to the parameters. The metric $\bM$ rescales movement by inverse curvature. The bias $\bb$ adds momentum or changes in the frame of reference. The noise $\bxi$ enables exploration. This framework unifies natural selection, Bayesian updating, Newton's method, stochastic gradient descent, stochastic Langevin dynamics, Adam optimization, and most other algorithms as special cases of the same underlying process. The Price equation also reveals why Fisher information, Kullback-Leibler divergence, and d'Alembert's principle arise naturally in learning dynamics. By exposing this common structure, the FMB law provides a principled foundation for understanding, comparing, and designing learning algorithms across disciplines.
}

\begin{document}

\mymaketitle

%% 1st parm is skip on left column at start of TOC, 2nd param is skip after TOC
\iftoc\mytoc{-24pt}{\newpage}\fi

\section{Introduction}

Learning algorithms pervade modern science. Machine learning improves neural networks through gradient descent. Evolution improves organisms through natural selection. Bayesian inference improves beliefs through probability updates. Despite decades of research in each field, the ultimate relations between these approaches remain unclear.

This article shows that a single force-metric-bias (FMB) law captures the essential structure of algorithmic learning and natural selection. Improvement arises from three components: force, typically expressed by the performance gradient; metric, typically expressed by inverse curvature; and bias, which includes momentum and other changes in the frame of reference. This structure emerges naturally from the Price equation, a simple notational description for the partitioning of change into components \autocite{price70selection,price72extension,frank12naturalb}.

Consider how the following two connections arise naturally within this framework. First, the primary equation in evolutionary biology \autocite{lande79quantitative}, $\GD\bth=\bP\mskip2mu\bGa$, and Newton's method \autocite{nocedal06numerical} in optimization, $\GD\bth=-\bHi\nabla U$, are mathematically analogous. Both describe one step of change by multiplying a gradient-like force, $\bGa$ or $\nabla U$, by evolution's covariance matrix, $\bP$, or Newton's inverse Hessian, $\bHi$, each serving the same metric role of rescaling geometry by inverse curvature.

Second, machine learning algorithms are used to improve the performance of neural networks or other methods of prediction. The progression between a few common algorithms perfectly illustrates the FMB decomposition. Stochastic gradient descent \autocite{robbins51a-stochastic} uses force, $\bff$. Polyak \autocite{polyak64some} adds momentum bias, $\bb$. Adam \autocite{kingma14adam:} includes adaptive metric scaling, $\bM$. Adam's full structure, $\bM\,\bff + \bb$, is the same as evolution's primary equation and Newton optimization, with an additional momentum bias term that often improves performance.

These connections reflect a deeper geometric principle. Many learning algorithms face the same fundamental challenge: maximize improvement in performance minus a cost paid for distance moved in the parameter space. Here, we must account for two aspects of geometry. First, there may be a curved relation between parameters and performance. Second, constraints on movement in the parameter space induce a metric that alters how distance is measured. For example, lack of genetic variation in a particular direction constrains movement in that direction.

The optimal solution is the product of the force and the inverse curvature metric. Different fields have discovered this same result within specific contexts. Here, we see it in its full simplicity and generality, providing a reason for the recurring role of Fisher information as a curvature metric in probability contexts and inverse Hessian metrics in local geometry contexts.

The geometric structure of learning has been partially recognized in prior work. Fisher \autocite{fisher25theory} and Rao \autocite{rao45information} established that statistical parameter spaces can have intrinsic curvature. Amari \autocite{amari98natural} used this insight to develop natural gradient methods.

In evolutionary biology, Shahshahani \autocite{shahshahani79a-new-mathematical} applied differential geometry to the dynamics of natural selection, introducing metric concepts to evolutionary theory. Newton's method uses curvature to improve stepwise updates. Machine learning algorithms are often designed to estimate local curvature in an efficient way. 

These insights about geometry, force, momentum, and bias remained confined to their domains. The full simplicity and universality of algorithmic learning have not been expressed in a clear and formal way. This article demonstrates the underlying unity, revealing the simple mathematical law that governs learning processes.

\section{The force-metric-bias law}

\subsection{Statement of the FMB law}

I first state the FMB law. The following subsection derives the law and clarifies the notation.

This law is not an empirical hypothesis but rather a universal mathematical structure that underlies learning or selection. The law is
\begin{equation}\label{eq:fmb}
  \GD\bthb = \bM\,\bff + \bb +\bxi.
\end{equation}
Here, $\bthb$ denotes a vector of $n$ mean parameter values that is updated by learning, optimization, or natural selection. The law also applies to updates of a single parameter vector, $\GD\bth$, instead of mean values. Here, \textit{parameters} are values of any sort. In biology, we call them \textit{traits}.

The $n\times n$ matrix $\bM$ describes a metric, which typically expresses the inverse curvature of the parameter space and the rescaling of distances. The nature of the metric varies in different algorithms, discussed below. Throughout this article, metric matrices that properly rescale distances are positive definite. When a matrix is not positive definite, algorithms typically modify it or use alternative metrics to ensure valid updates.

The force vector $\bff$ often includes the gradient $\nsub{\bth}\U$ of the performance function $\U$ with respect to the parameters $\bth$. In general, the force vector typically describes processes that push toward increased performance or that constrain such increase. 

The bias vector, $\bb$, includes processes such as parameter momentum or change in frame of reference. These processes alter parameters in addition to the standard directly acting forces imposed by performance or constraint. The standard form of bias is
\begin{equation}\label{eq:bias}
  \bb=\bC\,\bGb+\bGg,
\end{equation}
in which $\bC$ describes a bias metric, $\bGb$ is the slope of performance with respect to biased parameter changes, and $\bGg$ is the bias that is independent of performance. Most algorithms follow this pattern for bias, modifying specific terms according to particular learning goals.

The noise vector, $\bxi$, has a mean of zero. Commonly, we partition the noise into a metric term and a simple noise-generating process. For example, many algorithms use some variant of
\begin{equation*}
  \bxi=\sqrt{\bD}\,\bGe,
\end{equation*}
in which $\bD$ is a metric that reshapes the noise, and $\bGe$ is a basic noise process such as a Gaussian with a mean of zero and a standard deviation of one.

The following derivation of the FMB law reveals further key distinctions between directly acting forces, $\bff$, and bias, $\bb$. 

\subsection{Derivation from the Price equation}

The generality of the FMB law arises from simple notational descriptions of change. This subsection describes the key steps. Note that, at first glance, the definition of terms in the FMB law may not seem to match many common learning algorithms, such as stochastic gradient descent. Later sections make the connections.

A subsequent section shows that the same simple approach also leads to common methods and measures that frequently arise in learning algorithms, such as Fisher information, Kullback-Leibler divergence, and information geometry. This article ties these pieces together.

(1) We begin with the Price equation, a universal expression for change. A probability vector $\bq$ of length $m$ sums to one. Each $q_i$ weights an alternative parameter vector, $\bth_i=\bth_1,\dots,\bth_m$, with each parameter vector $\bth_i$ of length $n$. The symbol $\bth$ without subscript denotes the $m$-vector of alternative $\bth_i$, each parameter vector associated with a probability $q_i$.

To get started, assume that we have only one parameter, $n=1$, with $m$ variant values. Later, I show that the same approach works for $n>1$, with notation extended for vectors and matrices.

An update to the mean parameter value over the $m$ variants is $\GD\Gthb = \bq'\cdot\bth'-\bq\cdot\bth$, in which the dots denote inner products, and $\GD$ is the difference between an updated primed value and the original value. Rearranging yields the Price equation \autocite{frank12naturalb}
\begin{equation}\label{eq:price}
  \GD\Gthb = \dbq\cdot\bth \mskip4mu+\mskip4mu \bq'\mskip-3mu\cdot\GD\bth.
\end{equation}
This equation is simply the definition of change in mean value, rearranged into a chain rule analog for finite differences rather than infinitesimal derivatives. The first term is the change in frequencies holding the parameters constant. The second term is the change in parameter values holding frequencies at their fixed updated values.

(2) Define $w_i$ as the relative growth of the $i$th type, $q'_i=w_iq_i$, such that
\begin{equation}\label{eq:wdef}
  \GD q_i=q_i(w_i-1).
\end{equation}
In biology, $w_i$ is called the relative fitness of the $i$th type, describing how survival and reproduction alter the frequencies of the types, with $\wbar=\bq\cdot\bw=1$.

By the standard definition of covariance, $\GD\bq\cdot\bth=\cov(w,\Gth)$, and by the standard definition of expectation
\begin{equation*}
  \bq'\mskip-3mu\cdot\GD\bth=\sum_i q'_i\GD\Gth_i=\sum_i q_iw_i\GD\Gth_i=\E(w\GD\Gth).
\end{equation*}
With these definitions, the Price equation can be rewritten as \autocite{price72extension}
\begin{equation}\label{eq:priceC}
  \GD\Gthb = \cov(w,\Gth)+\E(w\GD\Gth).
\end{equation}
These forms of the Price equation are simply notational descriptions for change \autocite{frank12naturalb}. We have not assumed anything about the nature of the values or how they change. We have assumed that $w_i$ always describes actual frequency changes.

If the performance function, $\U$, subsumes all of the forces that act on frequency change in a particular time period, then $w_i$ is the performance of the $i$th type, $\U(\bth_i)$, normalized to $\wbar=1$ for notational simplicity and without loss of generality
\begin{equation*}
  w_i = \frac{\U(\bth_i)}{\sum_j\U(\bth_j)},
\end{equation*}
in which fitness and relative performance are equivalent descriptions of actual change.

In some cases, the forces acting on frequency change are composed of several distinct processes. One component may arise from a performance function, $\U$. Another component may act as a constraining force that prevents frequency changes from following the forces imposed by $\U$. Then frequency change is no longer aligned with the optimal direction for improving performance, and $w_i$ is not equivalent to the relative value of the performance function, $\U$. Nonetheless, $w_i$ is the actual relative performance in the context of the Price equation's notational conventions.

(3) To derive the first term of the FMB law in \Eq{fmb}, write the standard least-squares regression of fitness on parameter values as
\begin{equation*}
  w_i = f_{w\Gth}\,\Gth_i + \zeta_i,
\end{equation*}
in which $f$ is the regression coefficient of $w$ on $\Gth$, and $\zeta$ is the error uncorrelated with $\Gth$. Using this regression in the first Price covariance term yields
\begin{equation*}
  \df\bar{\Gth}=\cov(w,\Gth) = f_{w\Gth}\,\var(\Gth),
\end{equation*}
in which $\df$ denotes the partial change caused by the force, $\bff$, imposed by relative performance, $w$. In the multivariate case, with $n>1$ parameters, this same term expands to
\begin{equation}\label{eq:Mf}
  \df\bthb=\cov(w,\bth) = \bM\,\bff,
\end{equation}
in which $\bM$ is the covariance matrix of the parameters, defined by $\cov(\bth,\bth)$, and $\bff$ is the vector of partial regression coefficients for fitness, $w$, with respect to each of the $n$ parameters.

Here, $\df$ changes average parameter values only through changes in frequency. 

(4) Bias directly changes a parameter value. For a parameter influenced by bias, $\GD\Gth_i=\Gth'_i-\Gth_i\ne0$. To derive the bias term of the FMB law, write the regression of fitness on the changes in parameters
\begin{equation*}
  w_i = \Gb_{w\GD\Gth}\,\GD\Gth_i + \zeta_i.
\end{equation*}
Then the second Price term yields
\begin{equation*}
  \E(w\GD\Gth) = \cov(w,\GD\Gth) + \Gg = \Gb_{w\GD\Gth}\var(\GD\Gth) + \Gg,
\end{equation*}
in which $\Gg=\E(\GD\Gth)$. For $n>1$, the extended notation is
\begin{equation}\label{eq:b}
  \db\bthb=\E(w\GD\bth) = \bC\,\bGb + \bGg = \bb,
\end{equation}
in which $\bC=\cov(\GD\bth,\GD\bth)$ is the covariance matrix of $\GD\bth$, and $\bGb$ is the vector of partial regression coefficients for $w$ on $\GD\bth$. The symbol $\db$ denotes the partial change caused by bias.

(5) We add a noise term, $\dxi\bthb=\bxi$, to complete the FMB law. In the infinitesimal limit, the law has the standard form of a stochastic differential equation. The first two components, $\df$ and $\db$, define the deterministic drift change, and the third $\dxi$ component defines the stochastic diffusion change \autocite{rice08a-stochastic}.

(6) As the parameter distributions concentrate near their mean values, the variances and covariances become small. In the limit, we have updates to a single parameter vector, $\GD\bth$, and the regressions in the $\df$ and $\db$ terms converge to gradients. This limit recovers the common usage of gradients in learning algorithms that update single parameter vectors rather than updating mean parameter vectors over distributions. In this limit, the metrics given by the covariance matrices are replaced by other aspects of geometric curvature, as discussed in the following subsections.

At this point, the FMB law is simply a notational partition of change into specific parts. The value follows from the insight and unity this notation brings to the diverse and seemingly unconnected applications that arise in different studies of learning and natural selection.

\subsection{Metrics, sufficiency, and single-value updates}

The Price equation's force, bias, and noise terms in the FMB law of \Eq{fmb} can be expanded to
\begin{equation}\label{eq:priceM}
  \GD\bthb = \bM\,\bff + \lr{\bC\,\bGb+\bGg} +\sqrt{\bD}\,\bGe.
\end{equation}
For the change in the location of the parameter vector, $\GD\bthb$, the metrics $\bM$, $\bC$, and $\bD$, the regression-based gradients, $\bff$ and $\bGb$, and the additional bias component $\bGg$ are sufficient statistics to reconstruct the update. Additional information about frequencies, $\bq$, does not alter the change in the location of the parameter vector.

The sufficiency of the terms in \Eq{priceM} to describe the change in the location of the parameter vector is important. It means that the FMB law, although initially derived from the Price equation's population frequencies, also accurately describes changes to a single parameter vector.

The update depends only on the sufficient statistics, which are the metric matrices, the force vectors, and the intrinsic bias. In other words, we can invoke the common geometry that unifies updates to the mean vector, based on underlying frequencies of different parameter vectors, or updates to a single vector, based on alternative calculations of the sufficient statistics.

The Price equation's metric terms are covariance matrices. However, a covariance matrix is just a metric matrix. In a population interpretation, we call the matrix a covariance. In a geometric interpretation, we call the matrix a metric. Mathematically, they are equivalent.

Similarly, population regressions enter only as slopes that can equivalently be analyzed geometrically. Intrinsic bias can also arise from either a population or a purely geometric interpretation.

Natural selection and some learning algorithms build on population notions of frequency and mean locations. Many other learning algorithms build on single-value updates of metrics, gradients, and geometry. Both interpretations follow from the Price equation's FMB law. The difference arises in whether we assume that changing population frequencies set the metrics and slopes, or we assume that other attributes of a system set the geometry.

This conceptual shift allows the FMB law to unify disparate fields. As we will see, the metric $\bM$ in natural selection is the empirically observed covariance matrix of parameters. In Newton's method for optimization, the metric $\bM$ is the analytically calculated inverse Hessian matrix. The FMB law reveals that these are different choices for the metric in different contexts, all within the same underlying mathematical structure.

In the following sections, I first continue to emphasize the Price equation's population-based perspective. Later, I switch emphasis to single-value updates based purely on a geometric perspective. The two perspectives are different views of the same underlying FMB law.

\subsection{A spectrum of methods: from local to population}

In practice, the variety of algorithms forms a spectrum of information-gathering strategies. The spectrum runs across the spatial and temporal scope of the information they use to define the curvature metric, $\bM$, the force, $\bff$, and the bias, $\bb$. Here, spatial scope describes a population of parameter vectors considered at a point in time, whereas temporal scope describes a sequence of parameter vectors over time.

Two extremes define the spectral extent. At one side, the purely local methods obtain information for both metric and force from a single parameter vector. For example, Newton's method calculates the force vector as the first derivative of performance and the curvature metric as the inverse Hessian matrix, the second derivative of performance. Both derivatives are calculated with respect to a single parameter vector.

At the other side, purely population-based methods use a full spatial scope to define both a covariance metric and a regression-based force. In this case, curvature and force are averaged over a distribution of alternative parameter vectors. Here, I briefly mention a few classic examples to illustrate how various algorithms fall along this spectrum.

Amari's natural gradient is a hybrid method. It combines a purely local force, the gradient at a point, with a metric of extended spatial scope, the Fisher information metric of a distribution over alternative parameter vectors \autocite{amari98natural,amari00methods}.

Stochastic gradient descent samples a batch of local gradients. The average over the several precise local force vectors estimates the force for a population sample. In effect, the statistical sampling transforms the local gradient descent method into a quasi-population method \autocite{robbins51a-stochastic,goodfellow16deep,mandt17stochastic,bottou18optimization}.

Optimization methods like Adam substitute temporal scope for spatial scope. As the optimizer traverses the parameter space over time, it generates a historical sequence of parameter vectors. This sequence provides a population of parameter vectors over which the method combines the local gradients to estimate a momentum-like statistic that augments the local force and to build a diagonal metric that captures aspects of the spatially extended curvature metric \autocite{kingma14adam:}.

Later sections will develop these analyses in detail, showing how the variety of algorithms arises from particular information-gathering strategies and ways of calculating the components of the FMB law.

\subsection{Performance and cost functions: sign convention}

I set $\U$ as a performance function that provides increasing benefits as it rises in magnitude. The choice of a target function to maximize arose from the Price equation's biological convention of fitness as a beneficial attribute.

By contrast, many studies in numerical optimization and other fields take $\U$ as a cost function to be minimized. In this article, I adopt the maximization of $\U$ as the primary goal. Results for minimizing cost follow by substituting $-\U$ for $\U$. If this substitution is used to minimize cost, then the Hessian calculation for local curvature becomes the curvature of the cost function $-\U$, which inverts the sign of the Hessian used in maximization.

There is no difference except that one has to pay attention to the directions of change and the appropriate signs appended to terms.

\section{Natural selection, metrics, and curvature}

This section links the metric and force terms to natural selection, a topic that has a well developed theoretical foundation. The connection illustrates how the familiar concepts in biology associate with the more abstract geometric concepts of the FMB law. In this case, metric and force arise from the spatial extent notion of populations, the basis of the Price equation and the initial path to the general FMB geometry.

From \Eq{Mf}, an update caused solely by the first Price term for frequency change is
\begin{equation*}
  \df\bthb=\cov(w,\bth)=\bM\,\bff.
\end{equation*}
In biological studies of natural selection, this result is often called the Lande equation \autocite{lande79quantitative}. In that case, $\bth$ is a vector of $n$ trait values, $\bM$ is the covariance matrix of the trait values, and $\bff$ is the vector of partial regression coefficients of fitness, $w$, on trait values, $\bth$. The Introduction wrote the right side of this equation for biology as $\bP\mskip2mu\bGa$ to distinguish the biological terms. But now we use our standard notation, $\bM\,\bff$.

This classic equation of natural selection matches the primary update process used by most learning algorithms. The slope of performance (fitness) relative to the parameters (traits) creates the primary driving force for updates, $\bff$. The covariance matrix, $\bM$, defines the update metric. Other algorithms vary in the spatial and temporal extent used to calculate force and metric, the methods for the particular calculations, and supplemental bias and stochasticity components.

A metric changes the length and direction of the update path, $\df\bthb$, by modulating the forces acting in each direction of the parameter space. The expected gain in performance is the force, $\bff$, multiplied by the displacement, $\df\bthb$, yielding
\begin{equation}\label{eq:fMf}
  \E\lr{\df\wbar} = \bff\cdot\df\bthb=\bff^\top\bM\,\bff.
\end{equation}
A metric alters the sum of squares for a vector, changing its Euclidean length, with the requirement on $\bM$ that the resulting length be a nonnegative real value. We can drop the expectation when $\bff$ is calculated from an explicit performance function.

A metric has a natural interpretation in terms of inverse curvature. For example, if $\bM$ is a covariance matrix for $\bth$, then, along any direction, a small value implies that there is little variation in the values of the parameters in that direction and therefore relatively little opportunity to shift the mean of the parameters.

A probability distribution with a small variance is narrow and highly curved, linking large curvature to small variation. Thus, inverse curvature describes variance. The movement in a particular direction becomes the force in that direction multiplied by the variance or inverse curvature in that direction. High variance and a straight surface augment the force. Low variance and a curved surface deter the force. Many publications consider the geometry of evolutionary dynamics \autocite{shahshahani79a-new-mathematical,kirkpatrick09patterns,walsh18evolution}.

In biology, natural evolutionary processes set the covariance matrix as the update metric. In learning and optimization algorithms, one chooses the metric by assumption or by particular calculations from the data and the update dynamics. The way in which the metric is chosen defines a primary distinction between algorithms.

\begin{figure*}[t]
\centering
\includegraphics[width=0.7\hsize]{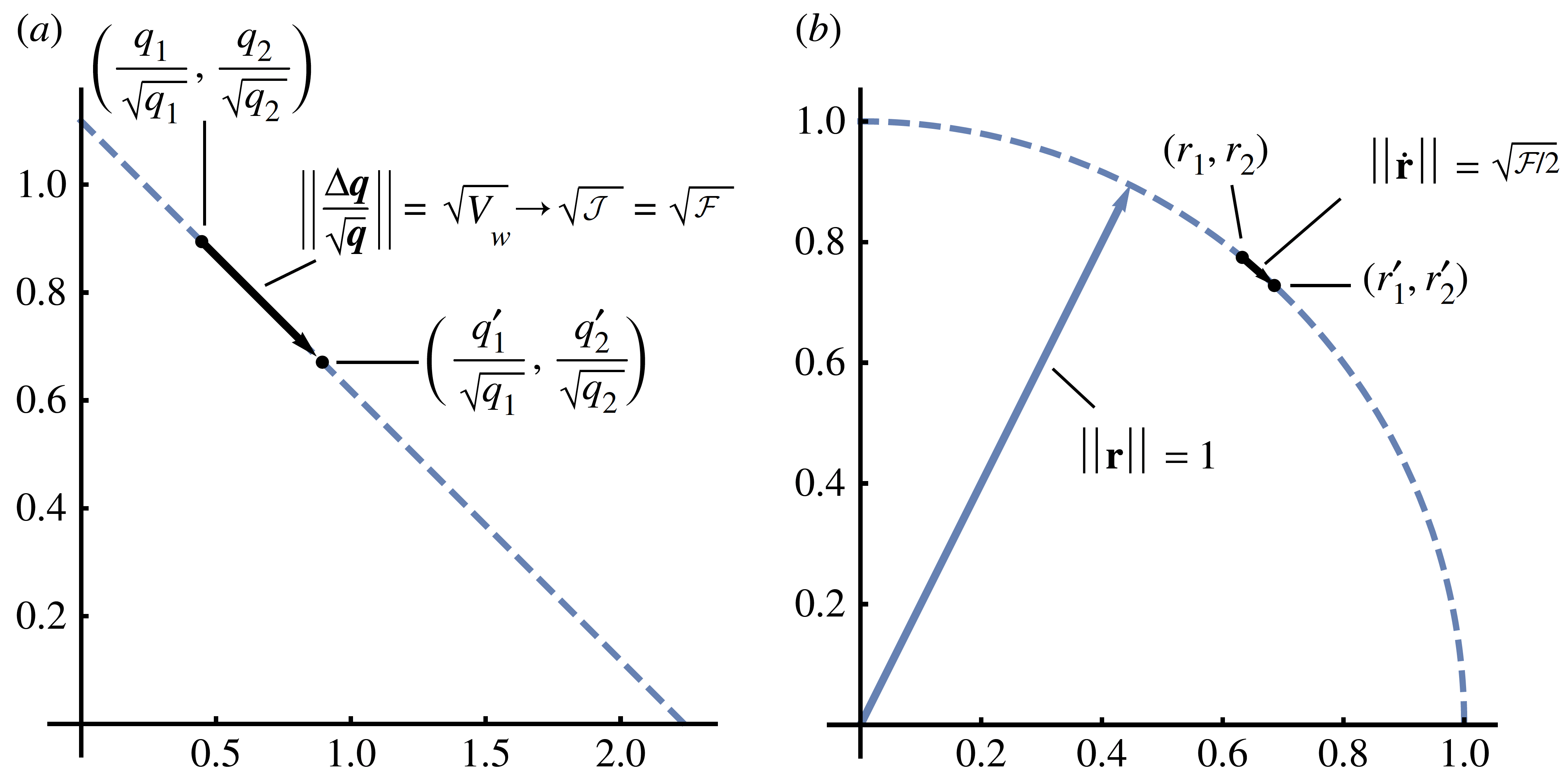}
\caption{Geometry of change by direct forces, $\df$. (\textbf{a}) Divergence between the initial population with probabilities, $\bq$, and the altered population with probabilities, $\bq'$. For discrete changes, the probabilities are normalized by the square root of the probabilities in the initial set. The distance can equivalently be described by the various expressions shown, in which $V_w$ is the variance in fitness from population biology, $\J$ is the Jeffreys divergence from information theory, and $\F$ is the squared Fisher-Rao step length. The symbol ``$\rightarrow$'' denotes the limit for small changes. (\textbf{b}) When changes are small, the same geometry and distances can be described more elegantly in unitary square root coordinates, $\br=\sqrt{\bq}$, which sets $\lVert\br\rVert=1$, and $\dot{\br}\equiv\dd\br=\dd\sqrt{\bq}=\left(\dd\bq\,/\sqrt{\bq}\right)/2$. From Frank \autocite{frank18the-price}.}
\label{fig:geometry}
\end{figure*}

\section{Geometry, information, and work}

Learning algorithms provide iterative improvement, a particular type of dynamics. This section reviews general properties of learning, which set a foundation for understanding the variety of algorithms and their unification \autocite{frank18the-price,frank20simple}.

We will see that the Price equation's notation for frequency change naturally gives rise to Fisher information, Kullback-Leibler divergence, information geometry, and d'Alembert's principle. The simple derivations reveal deep connections between learning dynamics and physical principles.

The Price equation and the consequent classic results follow from the intrinsic geometry of the purely population-based case. The insights from this spatially extended scope of populations provides the foundation to understand how the local geometric analysis of many learning algorithms fits within the broad FMB law.

\subsection{Price equation foundation}

The general expressions for learning updates arise from the Price equation, from which we see that the metric and force terms follow immediately from the basic notational description for the change in frequency.

Many learning algorithms do not have an intrinsic notion of frequency. Earlier, I showed how those algorithms fit into this scheme by considering the FMB terms as sufficient quantities for updates.

In this section, I continue to focus on frequency changes. Frequency here simply means a vector of positive weights with a conserved total value. We normalize the total to be one, which links to notions of probability, frequency, and average values. Several classic measures and methods of learning follow.

Most of the particular results in this section are widely known. Once again, the advantage here is that these aspects emerge simply and naturally as the outcome of our basic Price equation notation, without the need to invoke particular assumptions or interpretations.

\subsection{Discrete Fisher-Rao length}

A discrete generalization of the Fisher-Rao length follows immediately from \Eq{fMf}, which gave the partial increase in mean fitness, $\wbar$, as $\bff^\top\bM\,\bff$. We can express that same quantity purely in terms of frequencies by the first Price term from \Eq{price}. To do so, we use fitness as the trait of interest, $\Gth_i=w_i$, with $w_i=1+\GD q_i/q_i$ from \Eq{wdef}, yielding \autocite{frank18the-price}
\begin{equation}\label{eq:fisherinfo}
  \df\wbar=\dbq\cdot\bw = \sum_i\frac{\lr{\GD q_i}^2}{q_i}
  		=\norm{\frac{\dbq}{\sqrt{\bq}}}^2=\F.
\end{equation}
The notation $\norm{\cdot}$ denotes the vector norm, which is the Euclidean length of the vector, and $\F$ denotes the discrete generalization of the squared Fisher-Rao step length that arises from the Fisher information metric \autocite{rao45information}.

The value of $\F$ measures the divergence between probability distributions for the discrete jump $\dbq=\bq'-\bq$. Equivalently, $\dbq\cdot\bw=\cov(w,w)=\var(w)$, the variance in fitness, describes the same value.

\subsection{Kullback-Leibler divergence}

The Fisher-Rao length measures the separation between probability distributions. The Kullback-Leibler (KL) divergence provides another common way to measure that separation \autocite{kullback51on-information}. If we write $w_i=e^{m_i}$, so that the discrete update is $q_i'=q_i e^{m_i}$, then we can think of discrete change as a continuous path arising from the solution of an infinitesimal process that grows at a nondimensional rate proportional to $m_i$, which in biology is the Malthusian parameter. Thus
\begin{equation}\label{eq:mdef}
  m_i=\log\frac{q'_i}{q_i}=\log w_i,
\end{equation}
and using $\bmm$ instead of $\bw$ in the Price equation, the first Price term for the partial change of mean log fitness caused directly by $\dbq$ becomes
\begin{equation}\label{eq:jeffreys}
  \df\mbar=\dbq\cdot\bmm=\KL{\bq'}{\bq}+\KL{\bq}{\bq'}=\J,
\end{equation}
which is known as the Jeffreys divergence \autocite{jeffreys46an-invariant}, a symmetric form of the KL divergence of information theory
\begin{equation}\label{eq:kldef}
  \KL{\bq'}{\bq}=\sum_i q'_i\log\frac{q'_i}{q_i}.
\end{equation}
For infinitesimal changes $\dbq\rightarrow\ddbq$, we get
\begin{equation}\label{eq:mdq}
  \bmm\rightarrow\bw-\bm{1}=\frac{\dd\bq}{\bq}=\dd\log\bq,
\end{equation}
so that using $\bmm$ yields 
\begin{equation*}
  \prt_{\bff}\mbar=\ddbq\cdot\bmm=\ddbq\cdot\bw=\F,
\end{equation*}
showing that, for continuous infinitesimal changes $\df\rightarrow\prt_{\bff}$, the Jeffreys divergence for discrete changes based on $\bmm$ converges to the squared Fisher-Rao step length, $\J\rightarrow\F$.

\subsection{Information geometry}

Information geometry analyzes the distance between probability distributions on a manifold typically defined by the Fisher information metric. Simple intuition about information geometry follows if we transform to square-root coordinates for frequencies \autocite{amari00methods}.

Let $\br=\sqrt{\bq}$, which leads to $\norm{\br}=1$, creating unitary coordinates such that all changes in $\br$ lie on the surface of a sphere with a radius of one. In the new coordinates, the value of the squared Fisher-Rao step length in \Eq{fisherinfo} for the infinitesimal limit becomes $\F=4\norm{\dd\br}^2$. This surface manifold for dynamics illustrates the widespread use of information geometry when studying how probability distributions change. \Figure{geometry} shows aspects of the geometry.

\subsection{Bayesian updating}

The distinction between Bayesian prior and posterior distributions is another way to describe the separation between probability distributions. Following Bayesian tradition, denote $\Lt(\bD|\bth)$ as the likelihood of observing the data, $\bD$, given parameter values, $\bth$. To interpret the likelihood as a performance measure equivalent to relative fitness, $w$, the average value of the force must be one to satisfy the conservation of total probability. Thus, define
\begin{equation}\label{eq:likelihoodW}
  w_i=L_i=\frac{\Lt(\bD|\bth_i)}{\sum_i q_i\,\Lt(\bD|\bth_i)}.
\end{equation}

We can now write the classic expression for the Bayesian updating of a prior, $\bq$, driven by the performance associated with new data, $\bL$, to yield the posterior, $\bq'$, as $q'_i=L_iq_i$, or \autocite{harper09the-replicator}
\begin{equation}\label{eq:bayesUpdate}
  \bL=\frac{\bq'}{\bq}=\bw.
\end{equation}

By recognizing $\bL=\bw$, we can use all of the general results derived from the Price equation. For example, the Malthusian parameter of \Eq{mdef} relates to the log-likelihood as 
\begin{equation}\label{eq:logL}
  \bmm=\log\frac{\bq'}{\bq}=\GD\log\bq=\log\bL.
\end{equation}
We can then relate the changes in probability distributions described by the Jeffreys divergence (\Eq{jeffreys}) and the squared Fisher-Rao update length
\begin{equation}\label{eq:partialL}
  \df\mskip2mu\skew{-0.25}\bar{L} = \dbq\cdot\bL=\norm{\frac{\GD\bq}{\sqrt{\bq}}}^2=\F.
\end{equation}

\subsection{d'Alembert's principle}

We can think of the causes that separate probability distributions during an update as forces. Multiplying force and displacement yields a notion of work. Because we conserve the total weights as normalized probabilities, many learning updates require that virtual work vanishes for allowable displacements, yielding d'Alembert's principle \autocite{frank15dalemberts}.

From the definition for $\bmm$ in \Eq{mdef}, and for the infinitesimal limit in \Eq{mdq}, we have $\mbar=\bq\cdot\bmm=0$. By the chain rule for differentiation we can write a Price equation expression
\begin{equation*}
  \dd\mbar=\dd\bq\cdot\bmm+\bq\cdot\dd\bmm=0.
\end{equation*}
Using $\bmm=\dd\bq/\bq$ from \Eq{mdq}, noting that $\bq\cdot\dd\bmm=\dd\bq\cdot\dd\log\bmm$, and rearranging yields
\begin{equation}\label{eq:dalembert}
  \lr{\bmm+\dd\log\bmm}\cdot\Gd\bq=0,
\end{equation}
in which $\Gd\bq=\dd\bq$ is a small virtual displacement consistent with all constraints. This expression is a nondimensional form of d'Alembert's principle, in which the virtual work of the directly acting force for an update, $\bF=\bmm$, and displacement, $\Gd\bq$, is balanced by the virtual work of the inertial force, $\bI=\dd\log\bmm$, and displacement, yielding $\lr{\bF+\bI}\cdot\Gd\bq=0$.

In one dimension, we recover an analog of the familiar Newtonian form, $F=ma$, or $(F-ma)\Gd r=0$, showing that the force, $F$, has an equal and opposite inertial force, $ma$, for mass $m$ and acceleration, $a$. For multiple dimensions, we can rewrite \Eq{dalembert} in canonical coordinates and obtain a simple Hamiltonian expression \autocite{frank15dalemberts}.

The conservation of total probability often leads to a balance of direct and inertial components, expressed by the Price equation. For example, when we analyze normalized likelihoods such that the average value is one, $\skew{-0.25}\bar{L}=\bq\cdot\bL=1$, we have a conserved form of the Price equation for normalized likelihood
\begin{equation}\label{eq:canonicalL}
  \GD\skew{-0.25}\bar{L}=\GD\bq\cdot\bL+\bq'\cdot\GD\bL=0,
\end{equation}
in which the first term is the gain in performance for the direct force of the data in the likelihood, and the second term is a balancing inertial decay imposed by the rescaling of relative likelihood in each update. Notions of direct and inertial forces and total virtual work provide insight into certain types of learning updates, shown in later examples.

\section{Alternative perspectives of dynamics}

The preceding sections described the universal geometry of change revealed by the Price equation's notation. Fundamental concepts emerged naturally, including the Fisher-Rao length, information geometry, and Bayesian updating. In this context, the Price equation is a purely descriptive approach that reveals abstract, universal mathematical properties of learning updates.

However, in practice, learning dynamics are more than descriptions of change. Algorithms infer causes or deduce outcomes. This section links the Price equation's description to inductive and deductive perspectives.

By making these alternative perspectives explicit, we see why the same mathematical object, such as the Fisher information metric, arises as a simple notational consequence in one context, an empirical observation in another context, and a chosen design principle for optimal performance in a third context \autocite{jaynes03probability,pearl09causality}.

To clarify these perspectives, we begin with the fact that any dynamic process has three key components: an initial state, a rule for change, and a final state. Typically, we know or assume two of the components and infer the third. The following subsections consider the alternative perspectives in detail, connecting each back to the abstract Price equation foundation.

This section’s alternative perspectives of dynamics adds a complementary axis to the local‑to‑population spectrum of methods introduced earlier. I develop this dynamics axis at the population scale, providing the most general conceptual frame. That population context prepares the ground for later discussion of particular algorithms, many of which blend local geometry with broader spatial or temporal scope.

\subsection{Descriptive, inductive, and deductive perspectives}

In our Price formulation, the partial change associated with force, $\df$, describes the initial state as the frequencies, $\bq$, the rule for change as the fitnesses, $\bw$, and the updated state as $\bq'$.

(1) The Price formulation is a purely descriptive and exact expression because it tautologically defines the rule for change from the other two pieces, $\bw=\bq'/\bq$. Consequently, results that follow directly from the Price equation provide the general, abstract basis for understanding intrinsic principles and geometry \autocite{frank12naturalb}.

(2) In biology, actual frequencies change, $\bq\mapsto\bq'$. Those changes are driven by an interaction between the current state, $\bq$, and unknown natural forces. A biological system has, in effect, direct access to the data for initial and updated states but not for the hidden rules of change.

Natural selection implicitly runs an inductive process \autocite{frank20the-inductive,frank14the-inductive}. It infers aspects of the hidden rules for change, designing systems that use that inferred information to improve future performance. In general, a system may use data on the initial and updated states inductively to infer something about the hidden rules of change.

(3) Most mathematical theories and most learning algorithms run deductively. They start with the initial state and the rule for change and then deduce the updated state. For example, we might have $\bq$, and the performance function, $\bw=U(\bth)$, from which we calculate $\bq'$.

The updated state $\bq'$ is an intrinsic calculation or outcome of the system process. More commonly, in machine learning, the process acts on a single parameter vector, $\bth$, rather than as a population of alternative parameter vectors. Given $\bth$ and a performance function, the algorithm calculates an updated vector, $\bth'$.

In summary, dynamics has three components: initial state, rule for change, and final state. Descriptive systems define the rule from the other two, $\bw=\bq'/\bq$. Inductive systems start with the initial and final state and infer the rule, $(\bq,\bq')\mapsto\bw$. Deductive systems start with the initial state and the rule and deduce the final state, $(\bq,\bw)\mapsto\bq'$ for populations or $(\bth,U(\bth))\mapsto\bth'$ for single-vector updates.

\subsection{Fisher information in the three perspectives}

This subsection shows how to interpret the squared Fisher-Rao step, $\F$, in each of the three perspectives. In the pure Price equation, it follows simply and universally from tautological notation. In both the inductive and deductive cases, it is the optimal step in the sense that it maximizes the increase in performance relative to alternative steps of the same length.

\subsubsection{Descriptive perspective}

In the abstract mathematical perspective, use \Eq{wdef} to define the focal trait as the average excess in fitness
\begin{equation*}
  a_i = w_i-1=\frac{\GD q_i}{q_i}.
\end{equation*}
Then the partial change in fitness from the first Price term, from \Eq{fisherinfo}, is
\begin{equation*}
  \df\GD\wbar=\dbq\cdot\ba=\norm{\frac{\dbq}{\sqrt{\bq}}}^2=\F.
\end{equation*}
By analogy with \Eq{fMf}, this quantity can also be written as
\begin{equation*}
  \df\GD\wbar=\bff^\top\bM\,\bff=\ba^\top\bS^{-1}\,\ba,
\end{equation*}
here interpreted purely in frequency space such that the force is $\bff=\ba$, and the metric is $\bM=\bS^{-1}$, a matrix with entries $q_i$ along the diagonal.

The matrix $\bS$ with entries $1/q_i$ along the diagonal is the Fisher information metric for the probability distribution $\bq$, also called the Shahshahani Riemannian metric in certain applications \autocite{shahshahani79a-new-mathematical}.

Here, I use the Fisher metric in its geometric sense as the Shahshahani metric, a full-rank, diagonal matrix that defines the curvature of frequency space \autocite{shahshahani79a-new-mathematical,amari00methods}. In this geometric context, the inverse given here provides the length metric for the space of $\ba$ values. In classic statistical estimation theory, the Fisher matrix has a different interpretation that reduces its dimension and leads to a different inverse form \autocite{rao45information}.

In this pure Price case, the values of $\bff$ and $\bM$ arise directly from notation, without any additional concepts or assumptions derived from information or particular aspects of geometry.

Thus, the Fisher metric may arise so often in widely different applications because of its fundamental basis in simple definitions rather than in the more complex interpretations commonly discussed. Those extended interpretations are useful in particular contexts, as in the following paragraphs.

The point here concerns the genesis and understanding of fundamental quantities rather than their potential applications. For this pure Price case, $\bM$ arises purely from tautological notation and is related to the inverse Fisher metric.

\subsubsection{Inductive perspective}

In inductive applications, we begin with or observe $\bq\mapsto\bq'$. From those data, we may inductively estimate $\bff$, the slope of $w$ with respect to trait values, $\bth$. The given frequency changes also inductively improve the system's internal estimate for parameters that perform well by weighting more heavily the high performance parameter values.

In general, from \Eq{Mf}, the update is $\df\bthb=\bM\,\bff=\cov(w,\bth)$, and the system's partial improvement in fitness (performance) caused by frequency change is $\bff^\top\bM\,\bff$, which is the squared Fisher-Rao length. Here, $\bM$ is $\bth$'s covariance matrix, and $\bff$ is the vector of partial regression coefficients of $w$ with respect to $\bth$ or, in the infinitesimal limit, the inferred gradient of $w$ with respect to $\bth$.

The covariance matrix $\bM$ in parameter space, $\bth$, is related to the Fisher metric $\bS$ in probability space, $\bq$, by 
\begin{equation}\label{eq:JSJ}
  \bM=\bJ^\top\bS^{-1}\bJ,
\end{equation}
in which $\bJ$ is the matrix of parameter deviations from their mean values, $\bJ_{ij}=\Gth_{ij}-\bar{\Gth}_j$. This expression reveals the relation of the Fisher metric geometry for probabilities to the covariance metric geometry for parameters.

The Fisher–Rao update is optimal in the sense that, at each step, it maximizes the first-order performance gain minus a penalty proportional to the geometric length of the update. In particular, among all possible frequency changes, $\dbq$, that produce the same Fisher-Rao update length, the actual frequency changes for the given trait covariance matrix, $\bM$, lead to the greatest improvement in performance.

Equivalently, among all possible frequency changes, $\dbq$, that produce the same improvement in performance, $\df\wbar$, the actual frequency changes for the given trait covariance matrix, $\bM$, lead to the shortest Fisher-Rao length. In biology, these optimality results are trait-based analogs of Fisher's Fundamental Theorem for genetics \autocite{fisher58the-genetical,ewens89an-interpretation,ewens92an-optimizing,frank09natural}. 

Of course, forces other than intrinsic performance can alter frequencies. The more we know about those other forces, such as environmental shifts or directional mutation in biology, the more accurately we can account for the consequences of those forces and improve inductive inference about the causal relation between parameters and performance \autocite{lande83the-measurement,crespi89a-path-analytic,scheiner00using}.

\subsubsection{Deductive perspective: frequencies}

In deductive studies of frequency, we use $\bq$ and $\bw$ to calculate $\bq'$. Mathematically, there is nothing new here because $q'_i=q_iw_i$. However, this perspective differs because we take the $\bw$ as given and deduce $\bq'$. In other words, $\bw$ is an identified driving force, whereas in the inductive case, $\bw=\bq'/\bq$ is an observation about frequencies that leads to inference about the variety of forces that have acted to change frequency. However, because the mathematics is the same, we end up with the same covariance and other expressions for change.

\subsubsection{Deductive perspective: parameters}

In deductive studies of parameters, we use $\bff$ and $\bM$ to deduce system updates to parameter values. This approach becomes interesting when, instead of being constrained by the given frequencies, $\bq$, that set the covariance matrix of $\bth$ values as the $\bM$ metric, we instead choose $\bM$ to get a better increase in performance.

Suppose, for example, that for $\bff$ we use $\dd\U/\dd\bth$, the gradient of performance (fitness) with respect to the parameters. If the gradient provides the best opportunity for improvement in a particular direction of the parameter space but, among the given $\bq$ values, there is no parameter variation in that direction, then the associated covariance of $\bth$ and metric $\bM$ prevent the potential gain offered by the gradient.

In a deductive application, we may instead choose $\bM$ to take advantage of the potential increase provided by the gradient. As before, the parameter update is $\df\bth=\bM\,\bff$, and the gain in performance is $\df\U=\bff^\top\bM\,\bff$. In this case we use a local gradient so the steps are accurate to first order, we drop the bar over $\bth$ because we no longer have an underlying distribution, $\bq$, and we use $\U$ for performance.

An optimal update typically occurs when the metric $\bM$ is $\bGi$, in which $\bG$ is the Fisher information matrix in $\bth$ coordinates. That step is optimal in the sense that it provides the greatest increase in performance among all alternative updates with the same Fisher-Rao path length. For an optimal update, the squared Fisher-Rao length equals the gain in performance.

However, we require an arbitrary assumption to calculate the Fisher matrix. That matrix, and the associated optimality, depend on the positive weights assigned to alternative parameter vectors, which we usually express as probabilities. But, in this case, we are considering an update to a given vector, $\bth$, without any underlying variants associated with probabilities, $\bq$. So we must create a notion of alternative parameter values with varying weights.

We are free to choose those probability weights, $\bq$. A natural choice is to use Boltzmann probabilities of the performance function,
\begin{equation}\label{eq:boltzmann}
  \bq(\bth)\propto e^{bU(\bth)},
\end{equation}
in which $b$ is a constant value, the maximum of $U$ is not infinite, and $U$ is twice differentiable. Here, $b$ adjusts how quickly the log probabilities rise in proportion to performance, $U$.

For the parameter vector $\bth=\Gth_1,\dots,\Gth_n$, the $i$th row and $j$th column entries of the Fisher matrix are
\begin{equation*}
  \bG_{ij}=-\E\lr{\frac{\prt^2\log\bq(\bth)}{\prt\Gth_i\,\prt\Gth_j}\,\Bigg|\,\bth}.
\end{equation*}
The Boltzmann expression in \Eq{boltzmann} links the log-probabilities used in the Fisher matrix to the performance function, yielding
\begin{equation*}
  \bG_{ij}=-\E\lr{\frac{\prt^2 U(\bth)}{\prt\Gth_i\,\prt\Gth_j}\,\Bigg|\,\bth}
  			= -\E\lrb{\bH(\bth)},
\end{equation*}
Thus, the Boltzmann choice for probability weights means that the Fisher matrix is the negative expected value of $\bH$, the Hessian matrix of second derivatives of $U$ with respect to $\bth$. The expectation is over the Boltzmann probabilities for each Hessian evaluated at a particular $\bth$. Thus, when we choose our metric as $\bM=\bGi$, the inverse Fisher matrix, we are choosing a particular metric of inverse curvature.

In the inductive case, $\bM$ arises from the given frequencies for alternative parameter vectors. For this deductive case, we allow $\bM$ to vary and ask what matrix maximizes the gain in performance minus the cost for the Fisher-Rao path length. The next subsection shows that $\bM=\bGi$ is optimal in this context and, in general, that inverse curvature is often a good metric \autocite{amari98natural}.

\subsection{Why inverse curvature is a good metric}

Consider first the case in which the only information we have is the local gradient, $\bff$, and the local Hessian curvature, $\bH$, near a particular point, $\bth$, the local end of the spatial extent spectrum. Then a Taylor series expansion of performance at a nearby point up to second order is
\begin{equation}\label{eq:taylorU}
  \U(\bth+\GD\bth) = U(\bth) + \bff^\top\GD\bth + \GD\bth^\top\bH\GD\bth/2,
\end{equation}
in which $\bH$ is the Hessian matrix of second derivatives. If we consider a region of the performance surface in which all second derivatives are negative, then $\bMi=-\bH$ is a positive definite metric that describes local curvature. We can then write the gain in performance for a step $\GD\bth$ as
\begin{equation}\label{eq:taylorUM}
  \GD\U = \bff^\top\GD\bth - \GD\bth^\top\bMi\GD\bth/2.
\end{equation}
What step $\GD\bth$ maximizes the total performance gain up to second-order? Equivalently, what step maximizes the first-order gain, $\bff^\top\GD\bth$, for a fixed quadratic cost, $\GD\bth^\top\bMi\GD\bth=c$? Lagrangian maximization of the gain subject to the fixed cost yields the optimal direction for an update,
\begin{equation}\label{eq:newton}
  \GD\bth^*\propto\bM\,\bff,
\end{equation}
which in this context is the classic Newton optimization method. For maximization, the positive-definite metric is $\bM=(-\bH)^{-1}$, in which $\bH$ is the negative-definite Hessian matrix of the performance function $\U$. For minimization of a cost function, the sign of $\bH$ reverses, so that the metric becomes $\bM=\bHi$, and the force $\bff$ also flips its sign. Same idea, different signs \autocite{nocedal06numerical}.

In this case, the metric $\bM$ scales each component of the step inversely to the local curvature, pushing far in straight directions and contracting where the surface bends sharply.  That inverse-curvature rescaling gains the most in first-order performance change for a given quadratic cost.

Given that the optimal metric for a local step arises from the inverse Hessian, why use the more complex Fisher metric? One reason is that local optimality requires that the Hessian be negative definite for the maximization of performance or, equivalently, positive definite for the minimization of cost. Another reason is that a local calculation ignores the overall shape of the optimization surface. So a locally optimal step is not necessarily best with regard to broader goals of optimization.

By contrast, the Fisher metric is essentially an average of best local curvature metrics over a region of the optimization surface, weighting each location on the surface in proportion to a specified probability. This approach, known as the natural gradient, typically points in a better direction with regard to global optimization \autocite{amari98natural}.

In terms of our local-to-population spectrum of methods, the natural gradient combines the spatial population extent for the calculation of the curvature metric with the local extent for the analysis of the force gradient.

For the Boltzmann distribution, the probability weighting of a location rises with the performance associated with that location, emphasizing strongly those regions of the optimization surface associated with the highest performance. Thus, a Fisher step typically points in a better direction with regard to global optimization.

The Fisher metric also corrects common problems with local Hessians. For example, Hessians are not always proper positive metrics, whereas the Fisher matrix is a proper metric. In addition, local Hessians can change under coordinate transformation, whereas the Fisher metric is coordinate invariant. As the region over which the Fisher metric is defined converges to a local region, the Fisher metric converges to the local Hessian.

The optimality of the Fisher metric follows the same sort of Lagrangian maximization as for the local Hessian \autocite{amari00methods}. In particular, we maximize the same gain, $\bff^\top\GD\bth$, but this time subject to a fixed KL divergence, $\KL{\bq'}{\bq}$, between the probability weightings for variant parameter values, taken initially as $\bq(\bth)$ and after a parameter update as $\bq'=\bq(\bth+\GD\bth)$. Here, $\bq(\bth)$ is a general distribution that can take any consistent form. By the Taylor series, the KL divergence to second order is
\begin{equation*}
  \KL{\bq'}{\bq} = \GD\bth^\top\bG\GD\bth/2
\end{equation*}
for Fisher matrix $\bG$. We can use the right side in place of the quadratic term in \Eq{taylorUM}. Then the same Lagrangian approach yields the optimal update in the context of the Taylor series approximations as
\begin{equation}\label{eq:optfisher}
  \GD\bth^*\propto\bGi\bff,
\end{equation}
which has the same form as the classic Newton update in \Eq{newton} but with $\bM=\bGi$, the inverse Fisher metric in place of the positive definite form of the inverse Hessian. In the limit of small changes, twice the KL divergence becomes the squared Fisher-Rao path length, $2\D\rightarrow\F$. Thus, the optimality again becomes the maximum performance gain relative to a fixed Fisher-Rao length.

In practice, the Fisher metric does not always provide the best update step. That metric is based on a particular assumption about global probability weightings for alternative parameter vectors, whereas one might be more interested in the local geometry near a particular point in the parameter space. In some cases, a local estimate of the inverse curvature provides a better update or may be cheaper to calculate.

The variety of learning algorithms trade benefits gained for particular geometries against costs paid for specific calculations. However, inverse curvature remains a common theme across many algorithms.

\section{The variety of algorithms}

The following sections step through some common algorithms. The details show how each fits into the FMB scheme, how the various algorithms relate to each other, and why certain quantities, such as Fisher information, recur in seemingly different learning scenarios.

The key distinctions between algorithms arise from how each gathers information about components of the FMB law. The alternatives span the spectrum from local information taken at the current point in parameter space to spatially extended averaging over a population of alternative parameter vectors, and from current values to temporally extended averaging of past or anticipated future locations. This section provides a brief overview.

First, population and Bayesian methods represent the fully extended spatial scope end of the spectrum. These methods focus on changes in frequencies, $\dbq$. In biology, frequencies change between ancestor and descendant populations. In Bayes methods, frequencies change between prior and posterior distributions. The frequencies act as relative weights for alternative parameter vectors.

A particular algorithm can, of course, choose to modify how components of an update are calculated. However, populations set the foundation for analysis. Changes in parameter mean values, $\GD\bthb$, summarize updates. The metric $\bM$ typically arises from the covariance of alternative parameter values. Some methods use variational optimization and analogies with free energy, which links learning to various physical principles of dynamics \autocite{friston10the-free-energy,sakthivadivel22towards}.

Second, many algorithms update a single parameter vector. These methods fill in the other end of the spectrum and its middle ground. The purely local strategy, such as Newton's method, uses a metric and force gradient analyzed at a single point. Hybrid strategies choose metrics that incorporate a broader spatial scope, such as trust regions \autocite{conn00trust,conn09introduction} or the natural gradient \autocite{amari98natural}.

Third, all search methods trade off exploiting the directly available information in their spatial domain against exploring more widely to avoid getting stuck in local optima. Noise provides the simplest exploration method, often expanding the spatial scope of analysis. Broader scope can sometimes push the search beyond a local plateau to find the nearest advantageous gradient to climb.

Finally, modern optimizers often include temporal scope. Extensions to stochastic gradient descent, such as Adam \autocite{kingma14adam:} and Nesterov \autocite{nesterov83a-method,sutskever13on-the-importance}, explicitly use the history of updates to calculate a bias term, $\bb$. In these cases, bias explicitly applies physical notions of momentum, in which past movement of parameter values can push future updates beyond local traps on the performance surface.

Overall, common optimizers combine different spatial and temporal extents of gradient forces, inverse curvature metrics, bias, and algorithmic tricks that compensate for missing information, difficult calculations, and challenging search over complex performance surfaces. We see that the seemingly different algorithms all build on the same underlying principles revealed by the Price equation's FMB law.

I start with spatially extended population methods, which match most closely to the standard interpretation of the Price equation.

\section{Population-based methods}

In machine learning, one often improves performance by directly calculating the gradient of a performance surface. An updated parameter vector follows by moving along the gradient's direction of improved performance.

However, in many applications, one does not have a smooth differentiable function that accurately maps parameters to performance. Without the ability to calculate the gradient, one has to test each parameter combination in its environment to measure its performance. Such methods are called black-box optimization algorithms.

Covariance matrix algorithms (CMAs) often provide a good black-box optimization method \autocite{hansen10comparing}. These evolution strategy (ES) methods proceed by analogy with biological evolution. One starts with a target parameter vector and then samples the performance values for a set of different parameter combinations around the target. That set forms a population from which one can calculate an updated target parameter by an empirically calculated covariance matrix and performance gradient, as in \Eq{Mf}.

These methods have three challenges: choosing sample points around the current target, estimating the performance gradient, and calculating the new target from the covariance matrix and performance gradient. I briefly summarize how the popular CMA-ES method handles these three challenges \autocite{hansen01completely}. 

First, CMA-ES draws sample points from a multivariate Gaussian distribution. The mean is the current target parameter combination. The algorithm updates the covariance matrix to match its estimate of the performance surface curvature. The covariance is reduced along directions with high curvature and enhanced along directions with low curvature. 

The size of the parameter changes in any direction increases with the variance in that direction. Thus, the algorithm tends to explore more widely over straighter (low curvature) regions of the performance surface and to move more slowly in curved directions.

In biology, covariance decreases over time in directions of strong selection because better performing variants increase rapidly and reduce variation. That decline in variation degrades the ability of the system to continue moving in the same beneficial direction because distance moved in a direction is the selection gradient multiplied by the variance. Eventually, mutation or other processes may restore the variance, but it can take a while to restore variance after a bout of strong selection.

CMA-ES avoids that potential collapse in evolutionary response by algorithmically maintaining sufficient variance to provide the system with good potential to search the performance landscape. In essence, the algorithm attempts to choose the inverse curvature metric, $\bM$, to maximize the gain in performance, in which $\bM$ is the covariance matrix.

The algorithm chooses $\bM$ by modifying the inverse Fisher metric of its Gaussian sampling distribution, steadily blending in the weighted variation of the best-performing samples to estimate its covariance matrix for updates. 

The estimated covariance matrix typically converges to an approximation for the local inverse curvature metric. For performance maximization, this metric is given by $\lr{-\bH}^{-1}$ for the local Hessian, $\bH$, in which the negative sign ensures that the metric is positive definite. The Gaussian sampling process introduces stochasticity that corresponds to $\bxi$ in the FMB law.

For the second challenge, CMA-ES approximates a modified performance gradient by sampling candidate solutions, ranking them by fitness, and then averaging the parameter differences between the best candidates and the current mean. The average puts greater weight on the higher fitness candidates. By contrast, biological processes implicitly calculate the partial regression of fitness on parameters.

For the third challenge, the direction in which CMA-ES updates the target parameter vector is approximately the product of its internal estimates for the covariance matrix and modified performance gradient. That approximation follows biology's updating by $\df\bthb=\bM\,\bff$ given in \Eq{Mf}, which is the universal combination of force scaled by metric.

The same $\bM\,\bff$ structure occurs in other evolution strategy (ES) learning algorithms, which include natural evolution strategies \autocite{wierstra14natural}, large‑scale parallel ES \autocite{salimans17evolution}, and separable low‑rank CMA‑ES \autocite{akimoto20diagonal}. The next section moves to the local end of the spectrum, showing that the $\bM\,\bff$ structure remains when the population collapses to a single vector.

\section{Single-vector updates}

In population methods, we track a weighted set of alternative parameter vectors over a spatially extended region of the parameter space. We often summarize learning updates by changes in the population mean, $\df\bthb$. Here, I use the partial component, $\df$, of the FMB law to focus on methods that use only the force and metric terms.

Many common algorithms update a single local parameter vector, $\df\bth$, without tracking any variant parameter values. This section summarizes a few classic single-vector update methods. The next section adds noise to enhance exploration of complex optimization surfaces, as in the commonly used stochastic gradient descent algorithm. The following section adds bias, including a momentum term that often occurs in some common machine learning algorithms, such as Adam.

For the $\bM\,\bff$ component of the FMB law, the following algorithms differ mainly in the way that they choose $\bM$. In some cases, $\bM$ arises from the same local extent as the gradient force calculation. In other cases, an algorithm expands the temporal or spatial extent to obtain a broader calculation of the curvature metric or uses a mirror geometry to design specific curvature attributes into the method.

\subsection{Gradient descent: constant Euclidean metric}

The update is
\begin{equation*}
  \df\bth = \eta\mskip1mu\bff,
\end{equation*}
in which $\eta$ is step size, and $\bff$ is the gradient of the performance function with respect to the parameters, evaluated at the current parameter vector, a purely local method. The implicit metric is $\bM=\eta\bI$, in which the identity matrix, $\bI$, is the Euclidean metric with no curvature. This simple method typically traces a path along the performance surface from the current location to the nearest local optimum.

\subsection{Newton's method: exact local curvature}

In \Eq{newton} we noted that using the positive definite inverse Hessian for the metric, $\bM=\bHit$, yields Newton's method
\begin{equation*}
  \df\bth=\bHit\bff,
\end{equation*}
in which $\bHt=-\bH$ if we are maximizing performance, and $\bHt=\bH$ with $\bff\mapsto-\bff$ if we are minimizing cost.

The Hessian is the matrix of second derivatives of the performance function with respect to the parameters at the current single point in the parameter space, a purely local measure of curvature. We could also include a step size multiplier, $\eta$, for any algorithms. However, we drop that step size term to reduce notational complexity.

The inverse Hessian metric typically improves local optimization by orienting the direction of updates into an improved gain in performance relative to a cost paid for the squared Fisher-Rao path length movement.

On the downside, the second derivatives may not exist everywhere, the Hessian may be computationally expensive to calculate, the information about second derivatives may be lacking, and this method tends to get trapped at the nearest local optimum.

In theory, using the inverse Fisher metric in place of the inverse Hessian often provides a better update \autocite{amari98natural}. The Fisher metric measures the curvature of the performance function with respect to the parameters when averaged over a spatially extended region of the parameter space.

The section \textit{Why inverse curvature is a good metric} discussed the many theoretical benefits of Fisher information in this context \autocite{amari98natural,amari00methods}. For example, the global Fisher metric often reduces the tendency to get trapped at a local optimum when updating the parameters. However, in practice, one often uses local estimates of curvature to avoid the extra assumptions and complex calculations required to estimate the Fisher matrix.

\subsection{Quasi-Newton: temporal extension}

These methods replace the computationally costly exact Hessian calculation with simpler updates that accumulate curvature information over time.  In terms of our FMB analysis, $\bM$ is iteratively updated to approximate $\bHit$, so that the temporally extended estimation of the metric becomes part of the algorithm \autocite{nocedal06numerical}.

The Broyden–Fletcher–Goldfarb–Shanno (BFGS) algorithm and its variants are widely used \autocite{fletcher00practical}. After each step the method compares how the gradient changed to how the parameters moved. That comparison provides information about curvature.

From that curvature information, the algorithm keeps a running update of its estimate for the inverse Hessian matrix. Thus, the method estimates the curvature metric by using only its calculation of first derivatives and parameter updates, without ever directly calculating a second derivative or inverting a matrix.

\subsection{Trust regions: spatial extension}

Newton's local calculation of the curvature via the inverse Hessian is fragile. The local Hessian may not exist, it may change significantly over space, or it may not be invertible to provide the inverse curvature. Trust region methods may solve some of these problems by defining a local region of analysis \autocite{conn00trust,conn09introduction}.

One type of trust region method spatially extends the calculation of the curvature metric in a way that matches the FMB law's notion of a population. By combining a local force gradient with a metric that is a positive-definite average of the curvature over the region, the hybrid method follows Amari's natural gradient approach \autocite{amari98natural}. The spatial extension of the metric calculation often improves updates by providing more information about the shape of the local performance landscape.

\subsection{Mirror descent: transformational extent}

The previous examples in this section match the $\bM\,\bff$ force-metric pattern in a simple and direct way. By contrast, the mirror descent algorithm has a more complicated geometry that does not obviously map directly to our $\bM\,\bff$ pattern \autocite{nemirovskij83problem,beck03mirror}. However, the following derivations show that this algorithm does in fact use the same basic force-metric approach but with a more involved method to obtain the curvature metric. 

In this case, the challenge is that the curvature either cannot be calculated or is computationally too expensive to calculate. So one calculates curvature in an alternative mirror geometry obtained by transformation of the target geometry, in which the curvature has a simpler or more tractable form. Instead of improving the metric by spatial or temporal extent, one uses a transformational extent to a geometry that can be chosen for its anticipated benefits to algorithmic learning.

For example, in a Newton update, the inverse Hessian is taken with respect to the performance function, $\U(\bth)$. In mirror descent, one changes the geometry by choosing a strictly convex potential function $\varphi(\bth)$ that has a positive definite Hessian $\bH_\varphi$ over the search domain. The inverse $\bHi_\varphi$ provides a consistent positive-definite metric $\bM$ for the update, repairing any fragility of the Hessian of $\U$.

Assume throughout this subsection that we are maximizing performance despite the word \textit{descent} in the common name for this approach.

Allowing for variable step size, $\eta$, the generalization of the Newton update for maximization becomes
\begin{equation}\label{eq:mirrorNewton}
  \df\bth = \eta\bHi_\varphi\bff.
\end{equation}
This update is a first-order approximation of the mirror descent method \autocite{nemirovskij83problem,beck03mirror}. In effect, we are making a Newton-like step in an alternative mirror geometry with metric $\bHi_\varphi$, then mapping that step back to our original geometry for $\bth$.

This approach allows one to control the metric and to compensate for a geometry of the performance surface that does not have a simple or proper curvature. Because the change in mirror space is determined by the metric $\bHi_\varphi$, we end up with our simple force-metric expression from the FMB law.

The optimal update rule in \Eq{mirrorNewton} arises as the first-order approximation for the solution of the following optimization problem, which balances the performance gain in the target geometry against a distance penalty in the mirror geometry. The problem is
\begin{equation}\label{eq:mirrord}
  \bth_{t+1} = \arg\max_{\bth}\lrbr{\nabla\U(\bth_t)\cdot\lr{\bth-\bth_t} 
  		- \frac{1}{\eta}B_\varphi\lr{\bth||\bth_t}}.
\end{equation}
The new update vector $\bth_{t+1}$ maximizes the two bracketed terms. The first term is the first-order approximation for the total performance increase. The second term is the Bregman divergence, $B_\varphi$, between new and old parameter vectors measured in the mirror geometry defined by transformation, $\varphi(\bth)$. That divergence, defined below, is an easily calculated distance moved in the mirror geometry. Thus, we are maximizing the performance gain in the target geometry minus the distance moved in the mirror geometry scaled by $1/\eta$.

In some cases, the mirror transformation depends on local information that changes in each time step, denoted $\varphi_t$ to emphasize the time dependence. Here, for notational simplicity, we drop the $t$ subscript but allow such dependence.

In this update equation, the first right-hand term is the local slope of the performance gain relative to the parameter change, weighted by the amount of parameter change, yielding the total performance increase. In the second term, the Bregman divergence is
\begin{equation*}
  B_\varphi\lr{\bth||\bth_t} = \varphi(\bth) - \varphi(\bth_t) - 
  	\nabla\varphi(\bth_t)\cdot\lr{\bth-\bth_t}.
\end{equation*}

We can obtain the first-order approximation for the optimal update given in \Eq{mirrorNewton} by the following. Start by differentiating the terms in the brackets on the right-hand side of \Eq{mirrord}, evaluating at $\bth=\bth_{t+1}$, and setting to zero, which yields
\begin{equation}\label{eq:mirrorKKT}
  \nabla\varphi(\bth_{t+1}) = \nabla\varphi(\bth_t) + \eta\nabla\U(\bth_t).
\end{equation}
Noting that $\bth_{t+1}=\bth_t+\GD\bth$, a first-order Taylor expansion of $\nabla\varphi(\bth_{t+1})$ around $\bth_t$ is
\begin{equation*}
  \nabla\varphi(\bth_{t+1}) = \nabla\varphi(\bth_t) + \nabla^2\varphi(\bth_t)\GD\bth
  	+ \mathrm{O}\lr{\norm{\GD\bth}^2}.
\end{equation*}
Dropping the second-order error and substituting into \Eq{mirrorKKT} gives \Eq{mirrorNewton} to first order in $\eta$, noting that $\bH_\varphi=\nabla^2\varphi(\bth_t)$. Thus, even in this relatively complex case, the force-metric law is nearly exact for small updates. This method is purely local in the same sense as Newton's algorithm but uses a transformational extent to improve the analysis of the curvature metric.

\section{Stochastic exploration}

All gradient algorithms suffer from the tendency to get stuck at a local optimum. This section briefly summarizes two methods that use noise to escape local optima and explore the performance surface more broadly.

These methods match the FMB law, $\bM\,\bff+\bb+\bxi$, with no bias, $\bb$, and noise entering via the stochastic term, $\bxi$. The deterministic component of these methods, $\bM\,\bff$, is a single-value update driven by the gradient force, lacking a population.

During a search trajectory, when the magnitude of this deterministic gradient component is greater than the magnitude of the noise, the deterministic single-value update process dominates. When the gradient flattens or the step size weighting for the gradient shrinks, the noise dominates the updates.

In the noise-dominated regime, the temporal trajectory samples a population of parameter locations around the path that the deterministic trajectory would have traced. That temporal wandering changes the search from single-value updates to quasi-Bayesian population-based method, in which the population distribution is shaped by the covariance of the noise process \autocite{welling11bayesian,mandt17stochastic}.

Technically speaking, as the gradient to noise ratio declines, the temporal stochastic sampling effectively becomes an ergodic spatially extended population . Thus, these hybrid methods exploit the simplicity and efficiency of single-value updates when the gradient is strongly informative and exploit the broader exploratory benefits of populations when the step-size weighted gradient is relatively weak.

\subsection{Stochastic Langevin search}

This algorithm combines a deterministic gradient step and a noise fluctuation \autocite{welling11bayesian}. A simple stochastic differential equation describes the process
\begin{equation}\label{eq:stochdiff}
  \dd\bth=\nabla U(\bth)\dd t + \sqrt{2\bD}\,\dd \bmr{W}_t.
\end{equation}
The parameter vector update, $\dd\bth$, equals the gradient of the performance surface, $\nabla U=\bff$, with respect to the parameters, $\bth$, plus a Brownian motion vector, $\bmr{W}_t$, weighted by the square root of the diffusion coefficients in the matrix, $\bD$, that determines the scale of noise.

In practice, one typically updates by discrete steps, for example, at time $t$, the update may be
\begin{equation*}
  \GD\bth_t = \eta \bM\nabla U(\bth_t) + \sqrt{2\eta \bM}\,\bGe_t.
\end{equation*}
Here, $\eta$ adjusts step size. 

The metric $\bM$ scales motion in each direction, typically obtained by estimating the positive-definite inverse curvature, such as the inverse of the negative Hessian matrix. The vector $\bGe_t$ is Gaussian noise with mean zero and covariance given by the identity matrix. The overall noise, $\bxi_t=\sqrt{\bD}\,\bGe_t$, has a covariance matrix of $\bD=2\eta\bM$. Alternative discrete-step approximations for \Eq{stochdiff} may be used to improve accuracy or efficiency.

This update matches our FMB standard for learning algorithms, the gradient force multiplied by the inverse curvature metric. The noise term adds exploratory fluctuations in proportion to the inverse curvature metric, $\bM$.

When the gradient is relatively large compared to the noise, the deterministic force dominates. When the slope is flat, noise dominates. The weighting of noise by inverse curvature means that fluctuations explore more widely in directions with low gradient and small curvature. Such directions are the most likely to be associated with trend reversals, providing an opportunity to escape a region in which the gradient pushes in the wrong direction.

The relative dominance of the deterministic gradient component compared with the noise component can be adjusted by changing $\eta$, the step-size weighting of the gradient term. As the $\eta$-weighted deterministic component declines, the method increasingly shifts from single-value updates to population-based updates. Here, dominance by temporal noise creates a similar effect to a spatially extended population \autocite{welling11bayesian}, as described in the introduction to this section.

In theory, methods such as stochastic Langevin search provide an excellent balance between exploiting gradient information and exploring by noise. However, the algorithm can be very costly computationally for high dimensions and large data sets. Each step uses all of the data to calculate the gradient with respect to the high-dimensional parameter vector.  The size of the inverse curvature matrix is the square of the parameter vector length, which can be very large. Stochastic gradient descent provides similar benefits with lower computational cost.

\subsection{Stochastic gradient descent}

In data-based machine learning methods, stochastic sampling of the data creates exploratory noise, which often improves the learning algorithm's search efficacy. To explain, I briefly review the steps used by common gradient algorithms in machine learning.

The data have $N$ input-output observations. The model takes an input and predicts the output. The parameter vector $\bth$ influences the model's predictions. 

For each observed input, the model makes a prediction that is compared with the observed output. A function transforms the divergence between the model prediction and the observed output into a performance value. One typically averages the performance value over multiple observations. The gradient vector is the derivative of the average performance with respect to the parameter vector. 

In machine learning, the average performance is typically called the loss for that batch of data. In this article, we often use the negative loss as the positive performance value. Climbing the performance scale means descending the loss scale.

The total data set has $N$ observations. If one uses all of the data to calculate the gradient and then updates the parameters using that gradient, the update method is often called gradient descent.

This deterministic gradient descent method will often get trapped in a local minimum, failing to find parameters that produce better performance.

Instead of using all of the data to calculate the gradient, one can instead repeat the process by using many small random samples of the data. Data mini-batches lose gradient precision but gain exploratory noise \autocite{robbins51a-stochastic,goodfellow16deep,mandt17stochastic,bottou18optimization}.

The stochasticity from random sampling can help to escape local optima and find better parameter combinations \autocite{zhu19the-anisotropic}. Thus, the method is often called stochastic gradient descent. In this process, a parameter update is
\begin{equation*}
  \GD\bth = \eta\widehat{\nabla U}(\bth),
\end{equation*}
in which $\eta$ is a chosen step size weighting, and the hat over the gradient, $U(\bth)$, denotes an estimated value from the sample batch of data.

The estimated gradient is implicitly the true gradient plus sampling noise. Thus, in theory,
\begin{equation}\label{eq:sgd}
  \GD\bth = \eta\lrb{\nabla U(\bth) + \bxi},
\end{equation}
in which $\nabla U$ is the true gradient, and $\bxi$ is the sampling noise of the gradient estimate. The variance of the sampling noise scales inversely with the batch sample size. Choosing a good batch size plays an important role in the optimization process \autocite{keskar17on-large-batch,mccandlish18an-empirical}.

The deterministic component of parameter updates dominates when the noise is small relative to the gradient signal. The stochastic component dominates when the noise is large relative to the gradient signal. Larger batch sizes reduce the scale of noise relative to the gradient signal.

As noted in the prior subsection, noisy exploration provides the greatest benefit when both the gradient and the curvature are small. In that case, noise provides the opportunity to jump across a near-zero or mildly disadvantageous gradient to a new base from which the gradient leads to improving performance. The more a region curves in a disadvantageous direction, the more noise one needs to jump across it.

As the batch size declines, the method increasingly shifts from deterministic single-value updates to stochastically sampled population-based updates. Here, dominance by temporal noise creates a similar effect to a spatially extended population \autocite{mandt17stochastic,smith18a-bayesian}, as described in the introduction to this section.

Many algorithms build on stochastic gradient descent by adding inverse-curvature metric scaling and bias. The next section provides examples.

\section{Bias in modern optimization}

Several widely used machine learning methods include a bias term. The bias alters parameters in addition to the direct force of the performance gradient.

For example, a moving average of past parameter changes describes the update momentum. That momentum includes temporal information about the shape of the performance surface that goes beyond the information in the local gradient and curvature.

This section provides examples of bias in various machine learning algorithms. Before turning to those examples, I briefly review the role of bias within the Price equation and the FMB law.

\subsection{Brief review of bias}

The Price equation's full FMB law from \Eq{fmb} is
\begin{equation*}
    \GD\bthb = \bM\,\bff + \bb +\bxi.
\end{equation*}
The prior examples focused on the $\bM\,\bff$ direct force and $\bxi$ noise terms. This section considers the bias term of \Eq{bias}, repeated here
\begin{equation*}
    \bb=\bC\,\bGb+\bGg.
\end{equation*}
Bias describes deterministic changes in parameter values, $\GD\Gth_i=\Gth'_i-\Gth_i$, that are not caused by the directly acting forces, $\bff$. Here, we take $\bff$ as the regression or gradient of performance with respect to the parameters. 

Denote these bias changes as $\GD_{\bb}\bth$. Then $\bGg=\E_{\bq}(\GD_{\bb}\bth)$ describes the bias that is uncorrelated with the performance. The matrix $\bC$ is the covariance of the bias vector. The vector $\bGb$ is the regression of performance on the bias vector, which captures correlations between performance and bias.

For single-value updates to the location vector, we use $\bthb\mapsto\bth$. The components of bias lose their statistical meaning. Instead, $\bC$ is a metric for the space of bias vectors. The slope $\bGb$ is the gradient of performance with respect to the bias vector. The term $\bGg$ adds further bias.

To focus on bias in the FMB law, this section drops noise terms. In effect, assume very large batch sizes for single-value updates or very large population sizes for population mean updates. The following examples decompose particular update algorithms into their FMB components. 

\subsection{Prior bias: parameter regularization}

Consider the single-value parameter update
\begin{equation*}
  \GD\bth = \eta\lr{\nsub{\bth}\U-\Gl\bth}.
\end{equation*}
The metric is $\bM=\eta\bI$. The force is $\bff=\nsub{\bth}\U$. The bias terms are $\bC\,\bGb=0$ and $\bGg=-\eta\Gl\bth$. 

The gradient imposes a force that pushes parameters to improve performance. The bias term imposes a static force that pulls all parameters toward a prior value, in this case, the origin.

Those parameters that only weakly improve performance end up close to the prior. In practice, one often prunes the parameter vector by dropping all parameters that end up near the prior, a process called regularization \autocite{goodfellow16deep}.

\subsection{Momentum bias: Polyak}

Abbreviate the gradient at time $t$ as $\bg_t = \nsub{\bth}\U(\bth_t)$. Then we can calculate the exponential moving average of the gradient as
\begin{equation*}
  \bmm_t=(1-u)\bg_t + u \bmm_{t-1}.
\end{equation*}
A simple parameter update that includes history is \autocite{polyak64some}
\begin{equation*}
  \GD\bth_t = \eta \bmm_t = \eta(1-u)\bg_t + \eta u \bmm_{t-1}.
\end{equation*}
On the right side, the first term is a standard gradient term for the update, $\bff=\bg_t$, with a constant metric $\bM=\eta(1-u)\bI$. The second term is the bias caused by the momentum from past updates, $\bGg=\eta u \bmm_{t-1}$. In this case, the bias is not associated with performance, $\bC\,\bGb=0$.

The idea is that a strong historical tendency to move in a particular direction provides information about the performance surface that supplements the information in the local gradient at the current time. Thus, the algorithm uses the momentum from past updates to push the current update in the direction that has been favored in the past.

Roughly speaking, the method uses temporal extent to gain information about the shape of the performance surface curvature rather than using the spatial extent of populations. However, in this case, the curvature information is used to bias the update rather than to calculate the metric that rescales the local gradient.

\subsection{Momentum bias and metric: Adam}

The widely used Adam algorithm adds metric scaling to the basic momentum update  \autocite{kingma14adam:}. The metric arises from the exponential moving average of the squared gradient
\begin{equation*}
  \bv_t = (1-s)\bg^2_t + s \bv_{t-1}.
\end{equation*}
The match to FMB follows by 
\begin{align*}
  \bff&=\bg_t\\
  \bM&=\eta\lr{\sqrt{\bv_t}+c}^{-1}\\
  \bC&=\bM\\
  \bGb&=u\bmm_{t-1}/(1-u)=\bmm_t/(1-u)-\bg_t\\
  \bGg&=0
\end{align*}
for small constant $c$ in $\bM$. The update follows as
\begin{align*}
  \GD\bth_t &= \bM\,\bff + \bC\,\bGb\\
  			&=\frac{\eta\bg_t}{\sqrt{\bv_t}+c}
  				+\frac{\eta u\bmm_{t-1}/(1-u)}{\sqrt{\bv_t}+c}\\[5pt]
 			&=\frac{1}{\lr{\sqrt{\bv_t}+c}(1-u)}\lrb{\eta(1-u)\bg_t
  				+\eta u\bmm_{t-1}}\\[5pt]
  			&=\frac{1}{1-u}\lr{\frac{\eta\bmm_t}{\sqrt{\bv_t}+c}}
  				=\frac{1}{1-u}\,\bM\,\bmm_t.
\end{align*}
Note in the third line that the bracketed quantity on the right is the Polyak momentum update from the prior subsection. Thus, Adam is proportional to the Polyak update weighted by the metric $\bM$.

Here, the metric is based on the exponential moving average of the squared gradient, $\bv_t$. Instead of inverse curvature, this metric is an inverse combination of the magnitude of the gradient and the noise in the gradient estimate when using small random batch samples of the data. Thus, the metric reduces step size in directions that have some combination of large slope or high noise. 

In summary, the metric $\bM$, derived from the history of squared gradients, creates an adaptive learning rate tuned to the geometry of each direction, and the momentum bias $\bb$ provides a temporally smoothed, directed force.  Overall, the temporal extent provides an efficient method to estimate spatial aspects of geometry.

\section{Gaussian processes and Kalman filters}

This section returns to population-based algorithms in relation to force and metric. I first introduce Gaussian processes, a population-based Bayesian method that weights alternative functions by how well they describe data rather than weighting alternative parameter vectors. The Bayesian weighting of alternatives provides a natural measure of uncertainty \autocite{rasmussen06gaussian}. 

I then turn to the common Kalman filter method. That method is conceptually similar to Gaussian processes, using a time series sequence of learning updates to track a changing system rather than the typical one-shot learning update by which Gaussian processes describe static systems \autocite{kalman60a-new-approach,welch95an-introduction}.

In the context of our FMB law, the technical details of the calculations do not matter so much. Instead, the point of this section is that a Gaussian process is a spatially extended population algorithm that depends on our typical metric and force terms to learn about a static process. 

Similarly, a Kalman filter uses a Bayesian population approach but instead aims to track a dynamically changing system. The Kalman filter update also reduces to our basic metric and force terms, in which the spatial geometry of the metric is estimated by temporal updating, as in Adam.

\subsection{Gaussian processes}

Our previous population methods updated the individual probability weightings in $\bq$. Each probability weighting associates with a multivariate combination of parameter values.

In contrast to a parametric model, a Gaussian process looks for the best function rather than the best parameter vector. Suppose, for example, that we are studying air temperature in relation to various predictors, such as cloud cover and humidity. We measure the actual temperature, $y$, and the vector of predictors, $\bx$.

Previously, we had a set of parameters that told us how to link the predictors to the outcome. The relative weighting of each parameter combination, $\bq$, described a Bayesian distribution, which we updated from prior to posterior based on the data.

In a Gaussian process, we use a function, $g(\bx)$, to predict the temperature, with deviations $y-g(\bx)$ between observed and predicted values. The goal is to find the best function among the set of candidate functions, $\bg$, that describe the data.

Previously, we described the associations between different parameter values by a covariance matrix. In a continuous Gaussian process, we use a kernel function $k(\bx,\bxt)$ that tells us, for any two points $\bx$ and $\bxt$ in the domain of inputs, how to calculate $\cov\lrb{g(\bx),g(\bxt)}$. For example, a common kernel function is
\begin{equation*}
  k(\bx,\bxt) = \Gs_g^2\exp\lr{-\frac{\norm{\bx-\bxt}^2}{2\ell^2}},
\end{equation*}
in which $\Gs_g^2$ is the baseline variance when $\bx=\bxt$, and $\ell$ is the length scale for how quickly a function can change. Different kernel functions encode different assumptions about smoothness or periodicity, and their hyperparameters can be updated as new data arrive.

We do not explicitly specify the functions, $\bg$, and their associated probabilities, $\bq$. Instead, we assume that, for an input vector $\bx$, the average output we expect over all functions is given by the mean curve $\Gm(\bx)=\E\lrb{g(\bx)}$. The curve can be thought of as the plot of $\Gm(\bx)$ versus $\bx$.

Two outputs covary by the kernel function, $k(\bx,\bxt)$, which tells us how similar the functional output values at $\bx$ and $\bxt$ tend to be. For a particular set of inputs, $\bX=\lrbr{\bx_1,\dots,\bx_N}$, the covariance matrix $\bK$ has entries $k(\bx_i,\bx_j)$.

For inputs $\bX$, the associated vector of mean values and the covariance matrix define a particular multivariate Gaussian. That Gaussian is the Bayesian distribution over all candidate function values for these specific inputs.

The notion of stepwise updating arises because, with new inputs, we alter the mean function, $\Gm(\bx)$, and we may also choose to alter the kernel function. The new mean vector and covariance matrix define the updated multivariate Gaussian posterior distribution over functions.

To define the update, note that for the functions $\bg$, the prediction, $y$, for any particular input, $\bx$, is given by the mean function, $\Gm(\bx)$. The set of $N$ inputs generates the vector mean, $\gbar=\Gm(\bX)$, and we also have the associated vector of measured values, $\by=\lrb{y_1,\dots,y_N}$. The measured values include measurement error, $y_i=h(\bx_i)+\zeta$, in which $h(\bx_i)$ is the true value associated with input $\bx_i$, and $\zeta$ is the unbiased Gaussian measurement error with variance $\Gs^2$.

An update is $\GD\gbar=\Gm_1(\bX)-\Gm_0(\bX)$, in which $\Gm_0$ is the prior mean function, and $\Gm_1$ is the posterior mean function. Typically we set $\Gm_0$ by prior knowledge or assumption. In the update, the difference in mean functions defines $\Gm_1$, the posterior mean function by
\begin{equation}\label{eq:gpUpdate}
  \GD\gbar = \bM\,\bff, 
\end{equation}
our standard FMB form of the metric multiplied by the force, in which metric and force are
\begin{align*}
  \bM  & = \lr{\bK^{-1}+\Gs^{-2}\bI}^{-1}\\[3pt]
  \bff &= \Gs^{-2}\lr{\by-\Gm_0}.
\end{align*}
In the metric, $\bK$ is the covariance matrix of the Gaussian prior, and its inverse, $\bK^{-1}$, is the prior precision or, equivalently, the prior Fisher information matrix for the information in an observation about the mean parameter vector. The term $\Gs^{-2}\bI$ adds the likelihood Fisher information in the data, $\by$, with Gaussian noise $\Gs^2$.

Adding those two Fisher components gives the total Fisher information in the posterior distribution. For the Gaussian case here, the Fisher matrix is the Hessian of the negative log posterior, $-\log \bq(\bg|\by)$. Thus, $\bM$, the inverse of the total Fisher matrix, is the inverse curvature of the Bayesian log posterior probability weights for alternative functions with respect to variations in those functions.

The force is proportional to the deviation between observed values, $\by$, and predicted values by the prior, $\Gm_0(\bx)$. We scale that deviation by the information in the observed data, $\Gs^{-2}$, which is the inverse variance of the measurement noise.

\subsection{Kalman filters}

One typically uses a Gaussian process model to infer a static functional relation between inputs and outputs. A Kalman filter applies a similar approach to a dynamically changing system \autocite{sarkka13spatiotemporal}.

The following description of the Kalman filter takes a bit of notation. However, the point is simple. Suppose there exists a dynamically changing vector of hidden states. We know the basis for the hidden stochastic dynamics but cannot directly observe the system state.

Instead, we observe a correlated measure of the hidden values. From the observed measurements, we can repeatedly update the estimated mean vector of hidden values, in which those estimates have a Gaussian distribution of error.

The updates of the estimated mean have a simple metric-force expression. The metric is the covariance of the error distribution, which is the local inverse curvature in the space of estimated values. The force is the deviation between observed and predicted values on the measurement scale, weighted by the information in an observation relative to the hidden values.

Suppose, for example, that the state of some system, $\bx_t$, changes with time, $t$. One cannot directly observe $\bx_t$ but can measure a correlate, $\by_t$. We have an explicit stochastic dynamical system
\begin{align*}
  \bx_t &= \bF\bx_{t-1}+\zeta_t\\
  \by_t &= \bH\bx_t+\eta_t.
\end{align*}
Here, $\bF$ defines the deterministic dynamics of $\bx$, and the noise is $\zeta_t \sim \N\lr{\bmr{0},\bQ}$. One can observe $\by$, which is $\bx$ measured through the filter $\bH$ and subject to  $\N\lr{\bmr{0},\bR}$ measurement noise.

How can we use our observation of $\by_t$ to update our prediction for the underlying $\bx_t$ values?

A Kalman filter uses the data in each time step to update the Gaussian distribution $\N\lr{\bxh_t,\bP_t}$, in which the mean vector $\bxh_t$ is the current estimate for the underlying true values, $\bx_t$, and $\bP_t$ is the covariance matrix of the estimated values. A Bayesian update maps prior to posterior distribution, $\lr{\bxh_t^-,\bP_t^-} \mapsto \lr{\bxh_t^+,\bP_t^+}$, given the data, $\by_t$.

The updated prediction, $\GD\bxh_t=\bxh_t^+ - \bxh_t^-$, follows our standard FMB product of metric and force
\begin{equation*}
  \GD\bxh_t = \bM_t\bff_t.
\end{equation*}
The metric, $\bM_t=\bP_t^-$, is the prior covariance, which is the inverse of the local curvature of the estimate error. The prior covariance describes a population of plausible state trajectories at each time step.

The recursive temporal updating of this covariance metric, $\bP_t^{-}=\bF \bP_{t-1}^{+} \bF^{\top}+\bQ$, is similar to the way in which Adam uses temporal extent to estimate spatial aspects of geometry. However, the Kalman filter implements this principle through a formal dynamic model given by $\bF$, rather than by the purely empirical averaging of past gradients used by Adam. Thus, the different algorithms use temporal information in distinct ways to achieve the same type of spatial metric within the FMB law.

The force is
\begin{equation*}
  \bff_t = \bH^\top\bS_t^{-1}\bv_t.
\end{equation*}
Here, $\bv_t=\by_t - \bH\bxh_t^-$, is the difference between the observed values, $\by_t$, and the predicted values based on the prior mean, $\bxh_t^-$, scaled to the $\by$ coordinates by $\bH$. Thus, $\bv_t$ is proportional to the force for change expressed on the $\by$ scale. Multiplying by $\bH^\top$ rescales the force to the $\bx$ coordinates. Finally,
\begin{equation*}
  \bS_t = \bH\bP_t^-\bH^\top+\bR,
\end{equation*}
which is the covariance matrix of $\bv_t$. Thus, the inverse is how much Fisher information there is in an observed $\bv_t$ about the hidden $\bx_t$.

Overall, the Kalman filter's use of temporal extent to estimate the spatial geometry of the covariance metric provides an interesting contrast with the Gaussian process's typical one-shot purely spatial estimate of geometry.

\section{Bayesian learning}

Natural selection and learning are often interpreted in Bayesian terms. Bayesian-inspired methods also provide many important computational learning algorithms. This section summarizes some of those methods, placing them within our broad framework for learning. Once again, I emphasize the simple conceptual unity of seemingly different approaches in the study of learning.

\subsection{Brief review}

The first Price equation term links the change in probability distribution, $\dbq$, to the change in mean values, $\df\mskip2mu\bar{\bth}=\dbq\cdot\bth$. If we interpret the change in each frequency value, $q'_i = q_iw_i$, as driven by relative performance, $w_i$, then can think of $\dbq$ as the change in the probability distribution driven by the improvement caused by learning \autocite{harper09the-replicator,shalizi09dynamics,frank12naturalc,campbell16universal,czegel22bayes}.

The updating of probability distributions by learning matches the standard Bayesian update process. In \Eq{likelihoodW}, we equated relative likelihood with relative fitness, $L_i=w_i$. Thus, $q'_i=q_i L_i$, in which $i$ associates with the $\bth_i$ parameter vector, $q_i$ is the prior probability of the $i$th parameter vector, $L_i$ is the relative likelihood of that parameter vector given some data, and $q'_i$ is the posterior probability of the $i$th parameter vector. All of that fits exactly into our Price equation expressions for the gain in performance caused by natural selection or learning, leading to \Eq{partialL}, repeated here
\begin{equation*}
  \df\mskip2mu\skew{-0.5}\bar{L} = \dbq\cdot\bL=\norm{\frac{\GD\bq}{\sqrt{\bq}}}^2,
\end{equation*}
which shows that the partial increase in likelihood caused by the direct force of learning is the same enhancement of performance measured by the discrete squared Fisher-Rao path length that we see in many learning models.

We can also write
\begin{equation*}
  \df\bthb = \bM\,\bff
\end{equation*}
for the change between the posterior and prior distributions, in which $\bM$ is the covariance of $\bth$ over the prior, and $\bff$ is the slope of the relative likelihood with respect to the parameters. Thus, Bayesian updating is a standard metric-force update for a population model.

The following subsections provide details about the nature of forces, how to partition those forces into components and constraints, and how to link classic physical notions of force, work, variational optimization, and free energy to the FMB framework.

\subsection{Variational Bayes}

Bayesian updating is in theory a simple method. From the prior subsection, $q'_i=q_i L_i$, in which $L_i$ is the relative likelihood given in \Eq{likelihoodW}, repeated here
\begin{equation*}
  w_i=L_i=\frac{\Lt(\bD|\bth_i)}{\sum_i q_i\,\Lt(\bD|\bth_i)}.
\end{equation*}
The practical problem arises when it is difficult to calculate the sum in the denominator to get the proper normalizing value, which we need to make the total probability of the posterior equal to one. That sum can be difficult to calculate when each likelihood associates with a large number of parameters, or when we have many alternative parameter vectors to consider.

Suppose we define a performance function that improves as our currently estimated posterior moves toward the true posterior. Then the challenge matches a common learning problem, and we can apply standard learning algorithms.

The variational Bayes method uses this approach \autocite{blei17variational}. Let our candidate posterior probability distributions, $\bqhat(\bGf)$, be confined to a particular distributional form that depends on the parameters, $\bGf$. Then the problem becomes the search for $\bGf$ that minimizes the divergence of the assumed form for the posterior, $\bqhat(\bGf)$, from the true posterior for $\bth$ given the data, $\bq'(\bth|\bD)$.

We measure that difference between estimated and true posterior by the KL divergence, $\KL{\bqhat(\bGf)}{\bq'}$, defined in \Eq{kldef}. This minimization problem is a variational method because we are minimizing the divergence between the function $\bqhat(\bGf)$ and the true posterior, $\bq'$.

Variational Bayes methods \autocite{blei17variational} provide a common approach to search for $\bqhat(\bGf)$. The details are simple but require a bit of notation to make the steps clear. In addition, we will use notation that can plug easily into the Price equation in the next section. Before starting, I list some notational shortcuts, in which $\bD$ refers to data
\begin{alignat*}{2}
  q_i     &{}= q(\bth_i)   &\quad &\mathrm{prior\ distribution}\\
  \qhat_i &{}= \qhat(\bth_i;\bGf)   &\quad &\mathrm{estimated\ posterior\ distribution}\\
  q'_i 	&{}= q(\bth_i|\bD)   &\quad &\mathrm{true\ posterior\ distribution}\\
  \Lt_i	&{}= q(\bD|\bth_i)   &\quad &\mathrm{nonnormalized\ likelihood, \Eq{likelihoodW}}\\
  {}    &{}\mskip23mu q(\bth_i,\bD) {}&\quad &\mathrm{joint\ distribution}\ \bth\ \mathrm{and\ data}\\
  {}    &{}\mskip23mu p(\bD) {}   &\quad &\mathrm{probability\ of\ the\ data}\\
  {}    &{}\mskip23mu \E_{\bqhat} {}&\quad &\mathrm{expectation\ with\ respect\ to}\ \bqhat(\bGf).\\
\end{alignat*}

We derive the variational Bayes method by starting with the basic rule of conditional probability, $q(\bth_i|\bD)p(\bD)=q(\bth_i,\bD)$, which we write in our shorthand notation as
\begin{equation*}
  p(\bD) = \frac{q(\bth_i,\bD)}{q'_i} = \frac{q(\bth_i,\bD)}{\qhat_i}\frac{\qhat_i}{q'_i}.
\end{equation*}
Take the $\log$ of both sides and then the expectation over $\bqhat$, noting that $p(\bD)$ on the left is a constant given the data, and so the expectation drops out on that side
\begin{align}
  \log p(\bD) &= \E_{\bqhat}\lrb{\log \bq(\bth,\bD)} - \E_{\bqhat}\log\bqhat+\KL{\bqhat}{\bq'}\nonumber\\
  			&=\LL(\bGf) + \KL{\bqhat}{\bq'},\label{eq:logD}
\end{align}
in which 
\begin{equation}\label{eq:elboA}
  \LL(\bGf) = \E_{\bqhat}\lrb{\log \bq(\bth,\bD)} - \E_{\bqhat}\log\bqhat
\end{equation}
is called the evidence lower bound (ELBO).

The goal of variational Bayes is to minimize the divergence between the estimated posterior, $\bqhat$, and the true posterior, $\bq'$, measured by the KL divergence term in \Eq{logD}. Because the value of $\log p(\bD)$ is constant, and the KL divergence term is nonnegative, maximizing the ELBO, $\LL(\bGf)$, minimizes the divergence. Thus, variational Bayes targets the maximization of $\LL(\bGf)$.

We can rewrite the ELBO in a more convenient form by starting with the rule for conditional probability and the definition of likelihood
\begin{equation*}
  q(\bth_i,\bD) = q(\bD|\bth_i)q(\bth_i)=\Lt(\bD|\bth_i) q(\bth_i),
\end{equation*}
which on the right side is the product of the nonnormalized likelihood of the data given the parameters, $\Lt$, and the prior probability of the parameters, $q(\bth_i)$. Using this expansion in the expression for the ELBO in \Eq{elboA} yields an alternative form
\begin{equation}\label{eq:elboB}
  \LL(\bGf) = \E_{\bqhat}\lr{\log \Lt(\bD|\bth)} - \KL{\bqhat}{\bq},
\end{equation}
the expected log-likelihood taken over the estimated posterior, $\bqhat$, minus the KL divergence of the estimated posterior from the prior.

Intuitively, maximizing ELBO balances the gain from concentrating the updated probability on regions with the greatest log-likelihood versus the cost of departing from the prior. In other words, the prior sets the default from which one changes only by the weight of new evidence.

\subsection{Variational Bayes from the Price equation}

The Price equation elegantly describes how much the ELBO increases as the estimated posterior, $\bqhat$, departs from the prior, $\bq$.

The Price equation (\Eq{price}) can be rewritten with $\bq'\mapsto\hat{\bq}$ and $\bth\mapsto\bz$ as
\begin{equation*}
  \GD\zbar=\dbq\cdot\bz+\hat{\bq}\cdot\dbz.
\end{equation*}
Here, $\GD q_i = \qhat_i-q_i$ is the difference between the estimated posterior and the prior.

We are free to choose the trait values $\bz$. For $\GD\bth_i=\zhat_i-z_i$, let
\begin{align*}
  z_i &= \log\Lt_i - \log\frac{q_i}{q_i} = \log\Lt_i\\[4pt]
  \zhat_i &= \log\Lt_i - \log\frac{\qhat_i}{q_i},
\end{align*}
in which $\Lt_i=\Lt(\bD|\bth_i)$ is the likelihood for the $i$th parameter combination given the data. With these definitions, $\zbar=\LL(\bGf)$, the ELBO. Thus, the total change in the ELBO is
\begin{align*}
  \GD\zbar 	&=\sum\qhat_i\zhat_i-\sum q_i z_i\\
  			&= \lr{\E_{\bqhat}\lr{\log \bLt} - \KL{\bqhat}{\bq}}
  				- \lr{\E_{\bq}\lr{\log \bLt} - \KL{\bq}{\bq}}\\
  			&=\E_{\GD \bq}\lr{\log\bLt} - \KL{\bqhat}{\bq},
\end{align*}
which is $\GD\LL(\bGf)$, the difference between the ELBO at $\bqhat$ and the baseline ELBO value when the estimated posterior is equal to the prior, $\bqhat=\bq$, noting that $\KL{\bq}{\bq}=0$.

To analyze the two Price terms separately, we need
\begin{equation*}
  \sum\qhat_i z_i = \E_{\bqhat}\lr{\log\bLt}.
\end{equation*}
We can now write the first Price term as
\begin{equation*}
  \dbq\cdot\bz = \dbq\cdot\log\bLt = \E_{\dbq}\lr{\log\bLt},
\end{equation*}
which is consistent with our earlier interpretation in \Eq{canonicalL} of the likelihood as the direct force driving frequency changes. The second Price term is
\begin{equation*}
  \hat{\bq}\cdot\dbz=-\KL{\bqhat}{\bq}.
\end{equation*}
This term describes how the direct gains made by moving frequencies toward regions of high likelihood alter the frequency context, imposing a cost on performance in proportion to how far the frequencies have moved from the prior.

The idea is that the total change in the ELBO balances the direct gain in moving closer to the likelihood of the data against the changed-context cost of moving away from prior information. We can write that as
\begin{equation}\label{eq:delbo}
  \GD\LL(\bGf) = \dbq\cdot\log\bLt - \KL{\bqhat}{\bq},
\end{equation}
in which the first term describes the direct force of the data via the log-likelihood, $\log\bLt$, and the second term is a weighted average of the opposing inertial force imposed by the prior, $\log\bqhat/\bq$.

For an infinitesimal change in the posterior frequencies, $\Gd\bqhat$, using variational notation, $\Gd$, for small differences that are consistent with the conservation of total probability, the first variational derivative of the ELBO in \Eq{elboB} is
\begin{equation}\label{eq:Gdelbo}
  \Gd\LL(\bGf) = \lr{\log\bLt - \log\frac{\bqhat}{\bq}}\cdot\Gd\bqhat,
\end{equation}
which clearly separates the forces and the displacement. Integrating this variational derivative over the specific frequency changes from $\bq$ to $\bqhat$ yields the total change in \Eq{delbo}.

\subsection{Statics, constraints, and d'Alembert's principle}

In variational Bayes, we choose a family of probability distributions for the estimated posterior, $\bqhat(\bGf)$, that vary according to $\bGf$. For example, we might choose a multivariate Gaussian with uncorrelated dimensions and a covariance matrix with diagonal elements given by $\bGf$.

More complex distributions may improve the potential to increase the ELBO. But those more complex distributions may also make the computational search process for improving the ELBO more difficult.

Whatever the chosen distribution, any learning method that improves the parameter choice for $\bGf$ with respect to the performance measure $\LL(\bGf)$ can be used. The point in this subsection is not to consider the specific dynamics of learning but rather to place the forces that directly influence improvement by learning into a broader context of direct, inertial, and constraining forces.

In general, setting a distributional family for $\bqhat(\bGf)$ imposes a constraint on the learning process. The Price equation provides a simple way to connect that particular force of constraint to a broader understanding of forces and dynamics.

Analyzing a static system often provides a simple way to understand various forces. A static system occurs when the forces are in balance, causing the overall system to remain unchanged. Balanced forces provide a strong clue about how the individual components must be changing to conserve the overall system.

We obtain a conserved system in the Price equation by defining
\begin{align*}
  z_i &= \log\Lt_i\\[0pt]
  \zhat_i &= \bq\cdot\log\bLt,
\end{align*}
so that $-\GD z_i$ is the deviation between the force acting on each dimension of $\bq$ and the average force acting over all dimensions. Then, the two Price equation terms are
\begin{align*}
  \dbq\cdot\bz &= \dbq\cdot\log\bLt\\
  \hat{\bq}\cdot\dbz &= -\dbq\cdot\log\bLt 
\end{align*}
Recalling that $\bLt=\bq'/\bq$, we can expand the second term to include
\begin{equation*}
  \log\bLt = \log\frac{\bq'}{\bq} = \log\frac{\bqhat}{\bq} + \log\frac{\bq'}{\bqhat}.
\end{equation*}
If we consider small virtual displacements of the posterior, $\Gd\bqhat$, a variational form of the Price equation for changes in $\zbar$ becomes
\begin{equation}\label{eq:vbStatics}
  \Gd\zbar = \lr{\log\bLt - \log\frac{\bqhat}{\bq} - \log\frac{\bq'}{\bqhat}}\cdot\Gd\bqhat=0,
\end{equation}
recalling that all allowable deviations $\Gd\bqhat$ satisfy the conservation of total probability, so that the sum of all deviations is zero. This expression matches d'Alembert's principle from classical mechanics, illustrated by \Eq{dalembert}.

d'Alembert's principle emphasizes how various forces in a conserved, static system balance, whereas most analyses focus on the dynamics of the moving parts. In other words, d'Alembert changes emphasis to the forces rather than the moving bodies \autocite{lanczos86the-variational}. That perspective helps when the goal is to understand how a system works rather than in the calculation of the actual paths of motion.

In the d'Alembert context, we can parse the balancing forces in \Eq{vbStatics}. Let $\Gd\bqhat$ be a virtual displacement of the posterior frequencies consistent with all constraints of the system. Then $\Gd\bqhat\cdot\log\bLt$ is the virtual work done by the overall potential force, $\log\bLt$, acting to change frequencies. That overall potential force is opposed by the reactive inertial force of the prior, $\log\bqhat/\bq$, within the constrained space of allowable posterior distributions, $\bqhat$. The overall potential is also opposed by the residual potential, $\log\bq'/\bqhat$, a reactive inertial force for the imaginary movement from the constrained posterior, $\bqhat$, to the true posterior, $\bq'$.

Noting from \Eq{Gdelbo} how the ELBO changes with an infinitesimal update, we can write our static d'Alembert expression in \Eq{vbStatics} as
\begin{equation*}
  \Gd\LL(\bGf) - \Gd\bqhat\cdot\log\frac{\bq'}{\bqhat}=0.
\end{equation*}
The virtual work gained by improving the evidence lower bound is balanced by the reactive force from the remaining potential. If $\bq^*$ is the optimum within the constrained space, we can split the remaining potential into \begin{equation*}
  \log\frac{\bq'}{\bqhat}=\log\frac{\bq^*}{\bqhat}+\log\frac{\bq'}{\bq^*}, 
\end{equation*}
which is the remaining potential from the current posterior, $\bqhat$, to the constrained optimum, $\bq^*$, plus the potential from the true posterior, $\bq'$, to the constrained optimum.

Overall, the d'Alembert expression for the separation of balancing forces follows naturally from the Price equation description of a conserved system. We gain a clear sense of how the various forces influence the dynamics of learning.

\subsection{Friston's free energy models}

Friston built a unified brain theory on the principles of variational Bayes analysis. In essence, Friston sought a theory in which the minimization of a single quantity could unify three problems that had previously been treated as separate topics \autocite{friston06a-free,friston07variational,friston10the-free-energy}.

First, homeostatic maintenance of biological function requires opposing the inexorable entropic decay of order. Friston suggested that such maintenance arises from minimizing the long-term average surprise from environmental sensory input. In a Bayesian context, surprise is $-\log p(\bD)$, the negative log probability of the data.

Second, Bayesian analogies had been used for how brains perceive and learn. But Bayesian calculations are often complex and difficult. Perhaps brains use the easier variational Bayes algorithm to approximate Bayesian inference, which maximizes the evidence lower bound (ELBO) by descending the free energy gradient.

Third, theories of behavioral action often derive from optimal control or reinforcement learning. Those theories optimize some measure of reward or value. Friston showed that increased value typically associates with reduced surprise, providing a direct link to Bayesian methods and the equivalent free energy expressions.

This subsection shows how Friston's methods fit into our broader framework for algorithmic learning. Before starting, it is helpful to emphasize the underlying simplicity of the program.

In essence, all of Friston's analyses reduce to a simple prescription. Keep tuning your model so that it assigns higher probability to the data that you actually observe, while not making the model unnecessarily complicated.

The tuned model is the estimated posterior, $\bqhat$. The goal is to drive $\bqhat$ close to the true posterior, $\bq'$, which means reducing the divergence, $\KL{\bqhat}{\bq'}$. Friston's wording and technical steps sometimes make it difficult to keep that simplicity in clear focus.

Friston's approach defines free energy, $F$, in a learning context by starting with the surprise of the data, $-\log p(\bD)$. Using our previous expansion of that expression in \Eq{logD} and rearranging the terms yields
\begin{equation*}
  F(\bGf) = -\mathrm{ELBO} + \log p(\bD) = \KL{\bqhat}{\bq'},
\end{equation*}
in which $-\mathrm{ELBO}=-\LL(\bGf)$. Because $\log p(\bD)$ is a constant for a given data input, choosing the parameters, $\bGf$, to minimize the free energy is equivalent to maximizing the ELBO, as in standard variational Bayes methods.

It is easier to see that $F$ is a plausible analogy for free energy by using the alternative form for $\LL(\bGf)$ in \Eq{elboB}, yielding
\begin{equation*}
  F(\bqhat) = \KL{\bqhat}{\bq} - \E_{\bqhat} \lrb{\log\Lt(\bD|\bth)} + \log p(\bD)
\end{equation*}
in which I have written $F(\bqhat)$ in this case to emphasize that we can equivalently think of optimizing the estimated posterior, $\bqhat(\bth;\bGf)$, or optimizing the parameters, $\bGf$, that determine the estimated posterior.

The difference in free energy, $\GD F=F(\bqhat) - F(\bq)$ for a change $\dbq=\bqhat-\bq$, can be written from \Eq{delbo} as
\begin{equation}\label{eq:dG}
  \GD F = \KL{\bqhat}{\bq} - \dbq\cdot\log\bLt = -\GD\LL{(\bGf)},
\end{equation}
which, in Friston's language, is read as the tradeoff in free energy between the gain from increasing the accuracy, $\dbq\cdot\log\bLt$, and the loss from increasing the complexity, $\KL{\bqhat}{\bq}$. Here, complexity means pulling further away from the maximally disordered state defined by the prior.

Classically, free energy measures the direct force that increases order opposed by the intrinsic pull toward disorder. Similarly, our Price equation analysis also shows learning as the balanced improvement by direct forces against the opposing decay of performance caused by inertial forces. In my opinion, the Price equation analysis derives the same balance of forces more transparently than the free energy analogy.

With this technical background, I describe Friston's two main conclusions.

First, the brain is designed to behave as if it minimizes surprise. In practice, it selects the variational Bayes posterior that maximizes the ELBO, thereby reducing the free energy, $F$. With each update the new posterior becomes the next prior, $\bqhat\mapsto \bq$, and the corresponding surprise associated with the updated estimate for the probability of the data, $-\log \hat{p}(\bD)$, should be reduced.

Second, agents choose future actions that minimize expected free energy. To do that, they adopt an active inference policy that balances the tradeoff between exploitation and exploration \autocite{kaplan18planning}. For exploitation, choose actions with likely outcomes that match current preferences. For exploration, choose actions that provide information to improve preferences, increasing the value of future outcomes.

Friston embedded an action-inference process within his free energy framework. That approach provides a single quantity to minimize, in which minimization balances the gains from exploitation and exploration. By assuming a common metric within a Bayesian framing, one can develop testable hypotheses about how closely actual behavioral sequences match the predicted learning process.

Actions depend on the current behavioral policy, $\pi$. The free energy depends on policy
\begin{equation}\label{eq:action}
  F(\pi) = \mathrm{risk} - \mathrm{information\ gain},
\end{equation}
The exploitation component measures risk, the mismatch between preferred outcomes and actual outcomes. Friston measures risk by surprise, the mismatch between expected outcomes for a behavioral policy, $\pi$, and the actual outcomes. In probabilistic models, we write $-\log q(o|C)$ for the surprise of an outcome, $o$, given an expected outcome, $C$. We get the expected surprise by averaging over the frequency of the actual outcomes given the policy.

The exploration component measures the gain in information by linking the world's outcome state, $s$, to policy, $\pi$. The agent's prior belief is the probability distribution $q(s|\pi)$. After a round of behavior with actual outcomes, $o$, the agent has an updated posterior belief, $q(s|o,\pi)$. The KL divergence between the posterior and prior measures the gain in information. The negative value of that divergence is the free energy decrease for updating beliefs.

\subsection{Matching Friston to Fisher and d'Alembert}

As so often happens, we can link Friston's seemingly special results back to our common canonical forms for the Price equation, Fisher information, and d'Alembert's principle. Friston's choice of using logarithms and small changes means that we are considering small path updates, as in \Eq{dalembert}. That equation shows that the opposing direct and inertial forces typically lead to virtual work components that are equal and opposite Fisher-Rao path lengths.

In Friston's case, with $\hat{\bq}=\bq+\Gd\bqhat$ for small change $\Gd\bqhat$, and thus $\log \hat{\bq}/\bq\rightarrow\Gd\bqhat/\bq$, the KL divergence becomes
\begin{equation*}
  \KL{\bqhat}{\bq}\rightarrow \Gd\bqhat\cdot\log\frac{\bqhat}{\bq}
  =\norm{\frac{\Gd\bqhat}{\sqrt{\bq}}}^2.
\end{equation*}
Thus, noting that $\log\bLt=\log\bq'/\bq$, \Eq{dG} becomes
\begin{align*}
  -\Gd F	&=\lr{\log\frac{\mskip4mu\bq'}{\bq}-\log\frac{\bqhat}{\bq}}\cdot\Gd\bqhat\\[5pt]
  			&=\Gd\bqhat\cdot\log\frac{\bq'}{\bqhat},
\end{align*}
in which all displacements, $\Gd\bqhat$, are consistent with the constraint that the posterior remain within the space of allowable alternative probability distributions. 

If we split the direct force, $\log\bq'/\bq$, into the component between the current prior, $\bq$, and the constrained optimum, $\bq^*$, plus the component between the constrained optimum and the true posterior, $\bq'$, then as the updated prior approaches the local optimum, $\bq\rightarrow\bq^*$, we have
\begin{equation*}
  -\Gd F =\lr{\log\frac{\bq'}{\mskip5mu\bq^*}
  +\log\frac{\mskip7mu\bq^*}{\bq}-\log\frac{\bqhat}{\bq}}\cdot\Gd\bqhat,
\end{equation*}
in which the constraining force, $\log \bq'/\bq^*$ is orthogonal with respect to allowable displacements $\Gd\bqhat$ and drops out, yielding
\begin{equation*}
  -\Gd F = \Gd\bqhat\cdot\log\frac{\bq^*}{\bqhat}
  	=\lr{\log\frac{\mskip7mu\bq^*}{\bq}-\log\frac{\bqhat}{\bq}}\cdot\Gd\bqhat.
\end{equation*}
Also, for a prior near the constrained optimum and a small displacement that moves to the optimum, $\bq^*=\bq+\Gd\bqhat$, we have $\log\bq^*/\bq\rightarrow\log\bqhat/\bq$, yielding $\Gd F\rightarrow0$, with a local d'Alembert balance between the direct and inertial forces at the constrained optimum
\begin{equation*}
  -\Gd F=\lr{\log\bLt - \KL{\bqhat}{\bq}}\cdot\Gd\bqhat\rightarrow0.
\end{equation*}
Overall, Friston's free energy models of learning fit naturally within our Price equation framework.

\section{Hierarchical learning}

Biological populations have a natural learning hierarchy. Each individual inherits a set of parameters from its genes. Those parameters guide a learning process over the individual's lifetime.

Within an individual, learning changes an internal nongenetic parameter vector. That learning influences the individual's success in transmitting its genes to the future of the population. In this case, what an individual learns does not get transmitted. The global process influences only the genes that seed the initial state of each individual. Call this nonheritable learning.

In biology, Baldwin \autocite{baldwin96a-new-factor} was perhaps the first to recognize that such hierarchical separation can greatly accelerate the overall rate at which a system learns. Subsequent studies in biology have extended the idea that nonheritable developmental adjustments or learning by individuals explain many aspects of how populations have actually evolved over the history of life \autocite{west-eberhard03developmental}.

Alternatively, we may think of each individual or lower-level group as transmitting the parameter vector that it learns over time. Then the global process aggregates the learned parameter vectors from each lower-level unit, weighting each lower-level contribution by the relative performance associated with its transmitted vector. In biology, many studies of group selection emphasize the enhanced evolutionary potential that arises from hierarchical population structure \autocite{maynard-smith76group,wilson83the-group,queller92quantitative,west08social}. Call this heritable learning.

In both cases, the success of each lower-level unit in learning determines the fitness or performance of the unit. The distinction between the nonheritable and heritable cases concerns whether the transmitted parameter vector is the same as the initial seeding of the unit or is the updated vector produced by improvement through the unit's learning.

The two cases lead to different types of search. In nonheritable learning, only the initial seeding vector is transmitted, and the parameters evolve to encode a better learning process within lower-level units. In heritable learning, the final improved vector of each lower-level unit is transmitted, and the parameters evolve to encode a better reaction in direct response to the particular challenge.

Many computational algorithms exploit the potential benefits of hierarchical learning \cite{wang21meta-learning}. This section uses the Price equation and the FMB law to show how hierarchical learning fits into our simple and general framework for algorithmic learning and natural selection.

\subsection{Nonheritable learning in lower-level units}

In this case, we keep our standard FMB law. However, to account for lower-level learning within individuals, we evaluate the fitness or performance function, $w\equiv\U$, at the point
\begin{equation*}
  \btht = \bth + \GD_\Gt\bth,
\end{equation*}
in which $\bth$ is the initial parameter vector passed to the lower-level unit, and $\GD_\Gt\bth$ is the change in the parameter vector by learning within the lower-level unit over the time period $\Gt$. 

The global update, $\bthb=\bM\,\bff+\bb$, uses $\U(\btht)$ to calculate the force vector, $\bff$. We can also adjust the bias vector in response to lower-level learning.

In computational application, the parameter change caused by the internal learning process, $\GD_\Gt$, typically arises from a sequence of discrete learning updates based on a local algorithm. Usually, we can express the local algorithm in FMB form. For example, in the $i$th individual
\begin{equation*}
  \GD_\Gt\bth_i = \sum_{r=1}^\Gt \bM_i\bff_i+\bb_i,
\end{equation*}
in which the metric, force, and bias terms may change with context in each time step. Here, if we consider a population or Bayesian process within individuals, we would write $\GD_\Gt\bthb_i$ for the individual change.

As before, we can switch between a population interpretation in which FMB statistics come from the population attributes and a single-value parameter update interpretation in which we set FMB statistics by other methods. A similar equivalence continues at the global level.

This setup provides a good description of many biological scenarios. In those biological cases, genetics fixes the initial values of individuals and the heritably transmitted values. Internal learning affects performance and reproductive fitness but does not influence the transmitted parameters.

\subsection{Nonheritable learning: algorithmic examples}

Hinton \& Nowlan \autocite{hinton87how-learning} illustrated how individual learning transformed an unsolvable search into an easily solved one. Simplifying the original Hinton \& Nowlan model, assume that each inherited genotype is a bit string of length 20. A particular target string is set as success. All other strings have the same low value of performance.

A small population is initialized with random strings. The chance that any single string matches the target is roughly $10^{-6}$. No simple combination of mutation, recombination, and reproduction by strings is likely ever to match the target.

Each string was then allowed a learning period. From the initial string value, the bits were mutated randomly over several rounds. If one of the mutated strings matched the target, then that individual had higher fitness and contributed more copies of itself to the next generation. The transmitted copy is the initial seed, not the internally mutated version during learning.

In this scenario, fitness measures the probability that an initial string can be mutated into a target match. That probability increases as a string's divergence from the target declines. Thus, internal learning turned a performance function with all weight on a single point into a graded performance function that increases steadily as the seeded string approaches the target. Learning algorithms are very good at following an improving gradient. Thus, this example of a Baldwin learning process shows how simply and powerfully internal learning can improve search.

Newer methods extend the Baldwin approach. For example, Fernando et al. \autocite{fernando18meta-learning} evolved a population of neural networks to solve a particular challenge. Each network inherits a parameter vector that sets the initial conditions, the way performance is calculated, and the learning rate. Each network then learns through a fixed number of standard stochastic gradient descent rounds of updates.

The final state of the network determines its performance. However, only the initially inherited vector is used to seed the next generation, with the contribution of each vector weighted by its associated performance after local learning. In tests, the nonheritable learning within individuals improved performance.

These population models split into a nonheritable inner loop of learning within individuals and a heritable outer loop between individuals. The same split can be done with single-value parameter updates rather than populations. For example, model‑agnostic meta-learning improves a single vector of initial weights for a learning process \autocite{finn17model-agnostic}.

Those initial weights seed a task‑specific inner loop with multiple steps of stochastic gradient descent. After evaluation of the final state, the updated parameter weights from this inner loop are discarded. Only the gradient of the inner-loop's final performance with respect to the initial parameters is fed into the update process for the outer loop.

This update process for a single parameter vector provides the same Baldwin-type separation of the overall learning process as occurred in the population-based methods. Population methods are advantageous when a component of the learning process lacks an easily calculated performance gradient, and one has to use a statistical method to associate particular changes in parameter vectors with changes in performance.

The standard one‑level FMB law holds in all of these cases. The force $\bff$ in the outer loop is the regression or gradient of fitness on the seed parameters. The inner-loop learning dynamics influences the calculation of fitness for a given seed parameter vector. Otherwise, the parameter updates from the inner loop are discarded. 

The seed parameters of the outer loop often include hyperparameters, which are those parameters that control the inner-loop learning process rather than encode the parameters of the particular learned solution. Many other methods optimize hyperparameters and other attributes that control the architecture of the learning process in the outer loop and discard the parameter tuning in each inner-loop round \autocite{falkner18bohb:,liu19darts:,nichol18on-first-order,zoph17neural,hospedales21meta-learning,li24bridging}. 

\subsection{Heritable learning: recursive Price equation}

The prior methods used an inner loop to evaluate fitness but did not retain updated parameters from the inner loop. Other methods retain the parameter changes from the inner loop, forming a fully realized two-level learning process.

To analyze multilevel learning, we recursively expand the Price equation to describe a population hierarchy \autocite{hamilton75innate,frank98foundations,okasha06evolution}. This subsection shows the steps.

The expanded Price equation reveals the sufficient statistics for population change as a hierarchical FMB law. The following subsection links the hierarchical FMB sufficient statistics to learning algorithms that update a single parameter vector rather than a population of parameter vectors.

Our initial derivation of the Price equation in \Eq{price} set $\wbar=1$ and so dropped that term. For recursive expansion, it is helpful to keep the $\wbar$ term, restating the Price equation for the change in the mean parameter vector as
\begin{equation*}
  \wbar\GD\bthb=\cov(w,\bth)+\E(w\GD\bth).
\end{equation*}
This equation describes the change within a single population. Suppose we split the population into distinct groups indexed by $g$, and individuals within groups indexed by $j|g$.

The total change, which remains the same, can be partitioned into changes between groups and changes within groups. The first step is to write the same one-level Price equation expression with extra notation to emphasize our top-level hierarchy of groups
\begin{equation}\label{eq:priceg}
  \wbar\GD\bthb=\cov\lr{\wbg,\bthbg}+\E\lr{\wbg\GD\bthbg},
\end{equation}
in which the covariance and expectation are taken over $g$. Previously, we had a population of parameter vectors, $\bth$, each parameter vector associated with a fitness value, $w$. Now we have the same population of parameter vectors, but we call each vector the mean value of a group, $\bthbg$. Again, each parameter vector maps to a fitness value, which we now describe as the group mean fitness, $\wbg$. 

We expand recursively by noting that the expectation in the last term on the right includes $\wbg\GD\bthbg$, which is the same form as the left side of the equation but for the groups, $g$. Thus, for each group $g$, we can use the whole equation to expand recursively by writing
\begin{equation}\label{eq:pricei}
  \wbg\GD\bthbg = \cov(\wbgj,\bthbgj)+\E(\wbgj\GD\bthbgj),
\end{equation}
in which the covariance and expectation are taken over $j$ for fixed $g$. Here, we drop the overbars on the right side because, at the lowest level of our analysis, $j|g$, we can maintain ambiguity about the nature of the values $\wbgj$ and $\bthbgj$ with regard to whether or not they are averages over some lower level.

Substituting \Eq{pricei} into \Eq{priceg} yields a two-level recursive expansion of the Price equation. We can expand into any multilevel hierarchy, for example, groups, subgroups, individuals, parts with individuals, and so on. Once again, as long as we maintain consistent notation, the Price equation is just a tautologically true notational expansion.

\subsection{Hierarchical FMB law}

To develop the hierarchical FMB law, define
\begin{equation}\label{eq:fmbhierparts}
\begin{aligned}
  \bMB&=\cov(\bthbg,\bthbg)\\
  \bfB&=\reg(\wbg,\bthbg)\\
  \bMg&=\cov(\bthbgj,\bthbgj)\\
  \bfg&=\reg(\wbgj,\bthbgj)\\
  \bbg&=\E(\wbgj\GD\bthbgj),
\end{aligned}
\end{equation}
in which the notation $\reg(w,\bth)$ means the vector of partial regression coefficients of $w$ with respect to $\bth$. Here, subscript $B$ denotes between all groups, and subscript $g$ denotes within group $g$. In this notation, function arguments with $g$ subscripts denote average over $g$, and arguments with $j|g$ denote averaging over $j$.

A potential bias between groups is implicit in the within-group biases, $\bbg$, expressed by a constant component of the bias for each $j|g$, such that $\bbg=\bb_B+\bbgt$. Thus, we can write the total bias in the population as
\begin{equation*}
  \bb=\E\lr{\bbg} = \bb_B + \E\lr{\bbgt}.
\end{equation*}

The hierarchical FMB law follows by the same procedure used in \Eq{Mf} and \Eq{b}, extended here for hierarchical expansion
\begin{equation}\label{eq:fmbhier}
  \GD\bthb=\bMB\bfB \,+\, \bb_B \,+\, \E\lr{\bMg\bfg \,+\, \bbgt}.
\end{equation}
For simplicity, I drop the noise terms in this section. We can interpret this expression as a partition of our standard FMB law by noting that
\begin{align*}
  \bM &= \bMB + \E\lr{\bMg}\\
  \bff&= \bM^{-1}\lr{\bMB\bfB \,+\, \E\lr{\bMg\bfg}},
\end{align*}
in which the first line is the total covariance, and the second line follows directly from equating $\bM\,\bff$ with the parenthetical quantity on the right side of the second line, which is the total change by the product of metric and force taken over all levels. Thus, the hierarchical components add to the total FMB expression
\begin{equation*}
  \GD\bthb=\bM\,\bff\,+\,\bb.
\end{equation*}

In some cases, it may be useful to separate timescales explicitly between the hierarchical levels. For example, if we wish to track $K$ updates within groups for each between-group update, then we can rewrite \Eq{pricei} as
\begin{equation*}
  \wbg\GD\bthbg =\bMg\bfg+\bbg = \sum_{k=1}^K \bMg^{(k)}\bfg^{(k)}+\bbg^{(k)},
\end{equation*}
in which each term of the sum is the $k$th within-group learning update. The expressions in \Eq{fmbhierparts} subsume this expansion by defining terms with respect to initial values and final values over the course of a within-group process. However, in practice, additional factors such as noise may enter in each explicit update, making the detailed summation of steps a better guide for matching the recursive Price equation's broad conceptual framing to actual implementations.

\subsection{Multilevel selection: algorithmic examples}

Group selection has been widely discussed in biology \autocite{maynard-smith76group,wilson83the-group,queller92quantitative,west08social}, often using the Price equation's recursive expansion \autocite{hamilton75innate,frank98foundations,okasha06evolution}. Many algorithmic learning methods have exploited the potential benefits of splitting populations into a multilevel hierarchy. This subsection mentions a few examples.

Commonly used evolutionary algorithms in machine learning can be extended by dividing populations into groups. These methods explicitly base their approach on the biological analogy of group selection or multilevel selection \autocite{sobey18re-inspiring,chand21multi-level}.

Other population methods split learning into an outer loop and a heritable inner loop \autocite{jaderberg17population}. The multistep inner loops may run within single agents of the population. Occasionally, the current state of one or more of the inner-loop agents is copied to seed some of the agents for a new population. For example, a currently strong agent may be cloned to replace a weaker agent. This approach combines the broad exploratory benefits of populations with the efficient improvement benefits of gradient-based approaches running within individual agents \autocite{khadka18evolution-guided,khadka19collaborative}.

\subsection{Single-vector updates}

The Price equation's population analysis reveals the sufficient statistics for the updates to the mean parameter vector. In practice, instead of calculating those changing statistics for a particular population, we can choose those statistics by other assumptions or calculations. Our choice influences the learning trajectory's rate of gain, cost, and other attributes, allowing us to design learning algorithms to meet different objectives.

Substituting chosen or alternatively calculated values for the sufficient statistics of populations leads to updates of a single location parameter vector rather than a mean parameter vector. In the earlier subsection \textit{Metrics, sufficiency, and single-value updates}, I discussed how this transformation from population methods to single-value methods unifies natural selection, Bayesian methods, the variety of evolutionary population methods, and the common single-value algorithms.

In this section, we get the single-value form of the multilevel Price expansion by dropping the overbars for means, interpreting the previous mean vector as a single-value location descriptor, and choosing how we wish to calculate the sufficient statistics of the FMB expressions.

In this way, we can link the multilevel Price equation's population description to single-vector methods that combine outer and inner learning loops. Often, a hierarchical method of this sort will repeatedly run a fast inner loop by one learning algorithm, then occasionally pass the final result from the inner loop to a slow outer loop that uses a different algorithm. The outer loop then reseeds another round of the inner loop, achieving a separation of time scales. The following paragraphs list three examples.

First, the look-ahead optimizer uses for its fast inner loop any common algorithm, such as stochastic gradient descent or Adam. Multiple inner-loop steps explore the local performance surface. The updated parameters then pass back to the slow outer loop, which adjusts the previously stored parameters in the direction of the inner-loop update. The next round of the fast inner loop begins with newly adjusted parameters, repeating the cycle \autocite{zhang19lookahead}.

In FMB language, the inner-loop steps are the within-group update. The subsequent blending of the inner-loop result with the prior outer-loop parameters plays the role of between-group selection and transmission, reseeding the next round with an improved state.

Second, we can interpret stochastic weight averaging within our hierarchical framework. The method first trains a model with stochastic gradient descent, forming the outer loop. Then it runs a further sequence of training steps in its inner loop. The outer loop is then updated to a final value by averaging over samples of the inner loop parameters \autocite{izmailov18averaging}.

Roughly speaking, as the trajectory converges near a local optimum, the stochasticity of the inner-loop updates tends to sample more in flatter regions of the performance surface near the optimum rather than in narrow and sharp parts of the performance surface. The flatter regions associate with parameters that have less sensitivity in their performance and better generalization in their response to previously unseen inputs.

Typically, this algorithm uses one outer loop followed by one inner loop, and so is just a sequence of alternative training methods rather than a hierarchy. However, in challenging cases, the method could be extended to alternate hierarchically between outer-loop updates and tests of performance, followed by further inner-loop sampling and averaging as needed to improve performance.

Third, deep Q networks optimize behavioral sequences. To value an action within the behavioral sequence of a focal network, the algorithm combines the observed reward for that action plus a predicted reward for future actions. To calculate the predicted future reward, the algorithm uses as a target the behavioral sequence of another network with a fixed parameter vector \autocite{mnih15human-level}.

With that setup, the algorithm runs multiple updates of an inner loop that improves the performance of the focal network measured against the fixed target network. After a round of updates in the inner loop, the inner loop's parameter vector overwrites the target network parameters. In effect, the target network is slowly updated by an outer loop that copies the learned vector of the inner loop.

Put another way, each inner loop is a learning period for the target network. The target network is then updated by inheriting the learned parameters of the inner loop. In biology, this sequence describes cultural or Lamarckian inheritance of acquired traits.

These single-vector procedures separate time scales in a way that matches the hierarchical FMB structure. The fast inner learner refines performance. The slow outer updates select the particular learned refinement that seeds the next round. That pattern is similar to what happens in group selection, in which the fast timescale of within-group selection improves performance, and the slow timescale of between-group selection chooses which within-group improvements seed the next generation.

\section{Conclusions}

The Price equation reveals a universal mathematical structure of algorithmic learning and natural selection, the FMB law, $\GD\bth=\bM\,\bff+\bb+\bxi$. This simple decomposition unifies seemingly disparate approaches, from natural selection's primary equation to machine learning's Adam optimizer and from Bayesian inference to Newton's method.

Each algorithm represents a particular choice of how to calculate or estimate the metric $\bM$, the force $\bff$, the bias $\bb$, and any additional noise $\bxi$. Typically the metric describes inverse curvature, the force arises from the performance gradient, and the bias includes momentum, regularization of parameters, and changing frame of reference. In some cases, the metric encompasses broader notions of rescaling geometry, and the force comes from a different way of pushing toward improvement. But the essence of metric scaling and improving force remains the same.

The framework's power lies in its simplicity. By recognizing that many learning algorithms attempt to maximize the performance gain minus a cost paid for distance moved in the parameter space, we see why certain mathematical quantities recur across disciplines. For example, Fisher information emerges naturally as a curvature metric in population or probabilistic models, whereas estimates of the Hessian of performance with respect to parameters describe curvature in locally focused methods.

The widespread commonality of the FMB structure suggests that advances in one domain can inform others. Computational methods for estimating curvature may illuminate evolutionary dynamics. The similar structure of Kalman filters and common machine learning optimization methods may suggest new refinements. The simple form of hierarchical learning may extend hierarchy to standard methods.

Ultimately, the FMB law provides a principled foundation for understanding existing algorithms and for designing new ones. Most often, we are choosing among different implementations of the same underlying process.

\section*{Acknowledgments}

\noindent The Donald Bren Foundation and National Science Foundation grant DEB--2325755 support my research. 

%\vfill\eject

\mybiblio	% uses main.bib by default, add other bibs as needed

%\addcontentsline{toc}{section}{Appendix}
%\section*{Appendix}
%
%This section lists $\ldots$.

% used cuted package strip env to force balancing of columns
%\ifmulticol\begin{strip}\hbox{\null}\end{strip}\hbox{\null}\fi

\end{document}